\documentclass[runningheads]{llncs}
\usepackage{graphicx}
\usepackage{wrapfig}
\usepackage{tikz}
\usepackage{comment}
\usepackage{amsmath,amssymb} 
\usepackage{xspace}

\usepackage{enumitem}
\usepackage{color,soul}
\usepackage[accsupp]{axessibility}  
\usepackage{xcolor}
\usepackage{color, colortbl}

\usepackage{cite}
\usepackage{pifont}%

\usepackage{makecell}

\usepackage[utf8]{inputenc}        
\usepackage[T1]{fontenc}    
\usepackage{hyperref}       
\usepackage{url}            
\usepackage{booktabs}       
\usepackage{amsfonts}       
\usepackage{nicefrac}       
\usepackage{microtype}      
\usepackage{cleveref}
\usepackage{pdfpages}
\usepackage{float}
\usepackage{hhline}
\usepackage{array}
\usepackage{varwidth}
\usepackage{multirow}
\usepackage{makecell}
\usepackage{tabularx}
\usepackage{amsmath}
\usepackage[linesnumbered,ruled]{algorithm2e}
\usepackage{subfig}
\usepackage{graphicx}

\definecolor{mygreen}{RGB}{0 139 69}
\definecolor{mygreen2}{RGB}{0 205 0}
\definecolor{myred}{RGB}{205 38 38}
\definecolor{TartOrange}{HTML}{ff2e35}
\definecolor{Orange}{HTML}{ff7825}
\definecolor{Mango}{HTML}{ffc013}
\definecolor{AppleGreen}{HTML}{7cb81b}
\definecolor{Blue}{HTML}{1173b0}
\definecolor{BdazzledBlue}{HTML}{2e58a5}
\definecolor{Purple}{HTML}{5b3590}
\definecolor{Sunglow}{HTML}{FFCA3A}
\definecolor{Gray}{gray}{0.9}
\hypersetup{
	colorlinks=true,
	urlcolor=violet,
	citecolor=mygreen2,
}

\newcommand{\equ}[1]{Eq.~(#1)}

\newcommand{\cifar}{\textsc{Cifar-10}\xspace}

\newcommand{\cifarh}{\textsc{Cifar-100}\xspace}

\newcommand{\tinyimagenet}{\textsc{Tiny-ImageNet}\xspace}
\newcommand{\texture}{\textsc{Texture}\xspace}
\newcommand{\lsun}{\textsc{Lsun}\xspace}
\newcommand{\places}{\textsc{Places365}\xspace}
\newcommand{\svhn}{\textsc{Svhn}\xspace}

\newcommand{\sood}{\textsc{MoodCat}\xspace}
\newcommand{\G}{$\mathbf{G}$\xspace}
\newcommand{\E}{$\mathbf{E}$\xspace}
\newcommand{\D}{$\mathbf{D}$\xspace}

\newcommand{\osrcifar}{\textsc{CIFAR}\xspace}

\newcolumntype{Y}{>{\centering\arraybackslash}X}
\newcolumntype{s}{>{\hsize=.2\hsize}Y}
\newcolumntype{t}{>{\hsize=.5\hsize}Y}
\newcolumntype{u}{>{\hsize=.5\hsize}X}
\newcolumntype{b}{X}
\newcolumntype{P}[1]{>{\centering\arraybackslash}p{#1}}

\begin{document}
\pagestyle{headings}
\def\ECCVSubNumber{2088}  

\title{Out-of-Distribution Detection with Semantic Mismatch under Masking} 

\titlerunning{\sood for Out-of-Distribution Detection}
%
\author{Yijun Yang \and
Ruiyuan Gao \and
Qiang Xu}
\authorrunning{Y. YANG et al.}
%
\institute{
\underline{CU}hk \underline{RE}liable Computing Laboratory (CURE Lab.)\\
       Department of Computer Science and Engineering\\
       \textit{The Chinese University of Hong Kong}, Hong Kong S.A.R., China\\
        \email{\{yjyang, rygao, qxu\}@cse.cuhk.edu.hk}}

\maketitle

\begin{abstract}

This paper proposes a novel out-of-distribution (OOD) detection framework named \sood for image classifiers. \sood masks a random portion of the input image and uses a generative model to synthesize the masked image to a new image conditioned on the classification result. It then calculates the semantic difference between the original image and the synthesized one for OOD detection. Compared to existing solutions, \sood naturally learns the semantic information of the in-distribution data with the proposed mask and conditional synthesis strategy, which is critical to identify OODs. Experimental results
demonstrate that \sood outperforms state-of-the-art OOD detection solutions by a large margin.
Our code is available at \url{https://github.com/cure-lab/MOODCat}.
\keywords{OOD detection, Robust AI, Generative model}
\end{abstract}

\section{Introduction}
\label{sec:introudction}
Deep neural networks (DNNs) are trained under a ``close-world'' assumption~\cite{krizhevskyimagenet, he2015delving}, where all the samples fed to the model are assumed to follow a narrow semantic distribution. 
However, when deployed in the wild, the model is exposed to an ``open-world'' with all kinds of inputs not necessarily following this distribution~\cite{drummond2006open}.
Such out-of-distribution (OOD) samples with significantly different semantics may
mislead DNN models and generate wrong prediction results with extremely high confidence, thereby hindering DNN's deployment safety~\cite{nguyen2015deep, hendrycks2016baseline, hein2019relu, amodei2016concrete,dietterich2017steps}.

To distinguish OOD samples from the in-distribution (In-D) data, some propose to reuse the features extracted from the original DNN model to tell the difference\cite{hendrycks2016baseline,liang2018enhancing,EBO,wang2021can,lin2021mood,wang2021energy}. 
However, such a feature-sharing strategy inevitably results in the trade-off between the prediction accuracy for In-D samples and the OOD detection capabilities. 
There are also various density-based OOD detection methods~\cite{pidhorskyi2018generative,ren2019likelihood,choi2018waic}, which try to model the In-D data with probabilistic measures such as energy and likelihood. However, the trustworthiness of these measures is not guaranteed~\cite{kirichenko2020normalizing}.
Another popular OOD detection mechanism uses generative models (e.g., variational autoencoder (VAE)) to reconstruct the input~\cite{taylor2018improving,schlegl2017unsupervised}. Based on the assumption that In-D data can be well reconstructed while OODs cannot since they are not seen during training, one could measure the distance between the original input and the reconstructed one and detect OOD with a threshold. 
However, this assumption is not sound. There are cases where OODs are faithfully reconstructed with the generative models, causing misjudgements~\cite{kirichenko2020normalizing}. 

In this paper, we propose a novel distance-based OOD detection framework, named \emph{Masked OOD Catcher} (\sood), wherein we consider the semantic mismatch under masking as the distance metric. Specifically, for image classifiers, we first randomly mask a portion of the input image, use a generative model to synthesize the masked image to a new image conditioned on the classification result, and then calculate the semantic difference between the original image and the synthesized one for OOD detection. 

Our insight is that, the classification result carries discriminative semantic information and it imposes strong constraints onto the synthesis procedure, especially when trying to recover the masked portions. With \sood, for correctly classified In-D data, the generative model can use the unmasked region to make up the masked part with sufficient training. In contrast, for OOD samples that are semantically different, the synthesized image based on the classification result tends to be dramatically different, especially for the masked region. 

\sood is a standalone OOD detector, and it does not require fine-tuning the original classifier. Consequently, it can be combined with any classifier to equip it with OOD detection capability without affecting its accuracy.
We perform comprehensive evaluations on standard OOD detection benchmarks~\cite{yang2021semantically} with six datasets and four detection settings. Results show that our method can outperform  state-of-the-art (SOTA) solutions by a large margin. 
We summarize the contributions of this work in the following:
\begin{itemize}[noitemsep,topsep=0pt]
	\item We propose a novel OOD detection framework by identifying semantic mismatch under masking, \sood. To the best of our knowledge, this is the first work that \emph{explicitly} considers semantics information for OOD detection.
	\item We present a novel masking and conditional synthesis flow in \sood, and investigate various masking strategies and conditional generator designs for OOD detection.
	\item To tell the semantic difference between the original image and the synthesized one, we employ an anomalous scoring model composed of various quality assessment metrics (e.g., DISTS~\cite{ding2020image} and LPIPS~\cite{zhang2018unreasonable}) and a newly-proposed conditional binary classifier. 
\end{itemize}

The rest of the paper is constructed as follows.
Section~\ref{sec:related_works} surveys related OOD detection methods.
We detail our proposed \sood framework in Section~\ref{sec:overview}.
Section~\ref{sec:experiments} presents our experimental results and the corresponding ablation studies.
Finally, Section~\ref{sec:conclusion} concludes this paper.   
\section{Related Work}
\label{sec:related_works}

\begin{figure}[tb]
    \centering
    \includegraphics[width=0.8\linewidth]{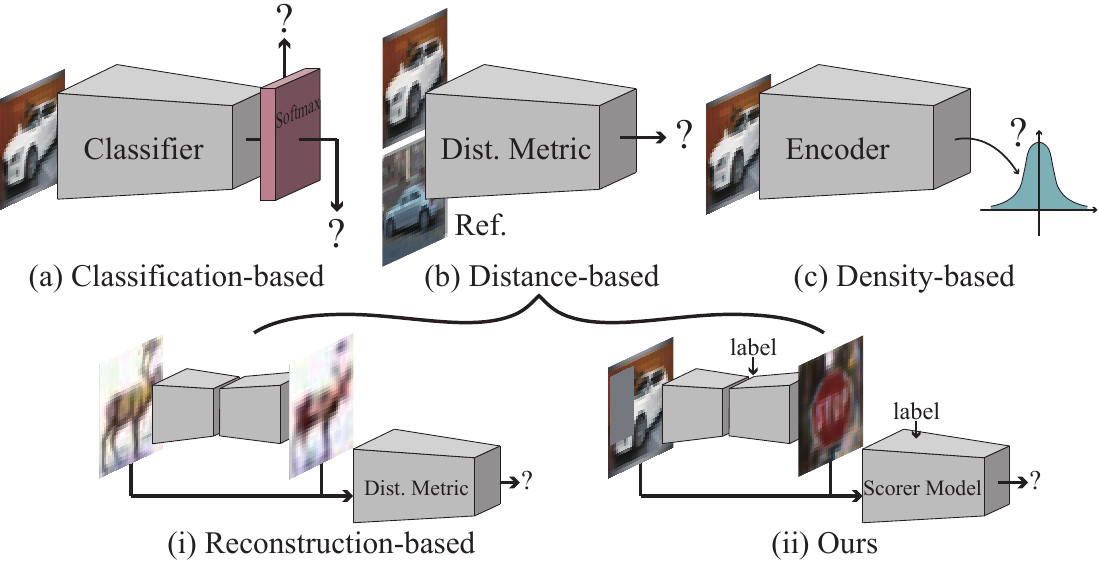}
    \caption{Comparison of OOD Detection Methods. \sood is a distance-based solution, and it relies on conditional image synthesis rather than reconstruction.}
    \label{fig:related}
\end{figure}

\subsection{Existing OOD Detection Methods}
In general, OOD detection methods can be categorized into: classification-based, density-based and distance-based methods~\cite{yang2021generalized}.

Classification-based methods derive OOD scores based on the output of DNNs, as shown in Fig.~\ref{fig:related}(a).
Maximum Softmax Probability (MSP)~\cite{hendrycks2016baseline} simply uses the maximum softmax probability as the indicator of In-D data.
ODIN\cite{liang2018enhancing} is applies a temperature scaling to the softmax value for OOD detection.
Follow-up works include methods based on the output of DNNs~\cite{EBO,wang2021can,lin2021mood, wang2021energy}, the gradient of DNNs~\cite{huang2021importance} and data generation or augmentation\cite{vernekar2019out,sricharan2018building}.
Although simple to implement, most of them alter the training process of the original classifier, thereby reducing the classification accuracy for In-D samples.

Density-based methods usually apply some probabilistic models for the distribution of In-D samples and regard test data in low-density regions as OOD~\cite{lee2018simple,ren2019likelihood,choi2018waic}, as shown in Fig.~\ref{fig:related}(c).
Some methods in this category also resort to generative models~\cite{pidhorskyi2018generative} to learn the distribution of data.
However, recent research found that the learned density model may assign high likelihood value to some OODs, since the obtained likelihood could be dominated by low-level features such as location and variance instead of the high-level semantics, which is related to the specific network architecture and data used for learning~\cite{nalisnick2018deep, choi2018waic}. 

Distance-based methods consider that OODs should be relatively far away from In-Ds. They either calculate the centroids of In-D classes in the feature space~\cite{zaeemzadeh2021out, huang2020feature} (Fig.~\ref{fig:related}(b)) or reconstruct the input itself (Section.~\ref{sec:2.2 reconstruction}, Fig.~\ref{fig:related}(i)) for OOD detection.
However, for high-level semantic features, their assumption for distance disparity may not hold, and high reconstruction quality cannot ensure In-Ds.
In this paper, we use conditional synthesis on masked images to highlight the semantics difference in the image space.


\subsection{Reconstruction-based OOD Detection}
\label{sec:2.2 reconstruction}
Reconstruction-based methods, which fall into the category of distance-based methods, are closely related to the proposed \sood technique.
These methods are based on the assumption that In-D data can be well-reconstructed from a trained generative model, but OOD cannot as they are not seen during training (see  Fig.~\ref{fig:related}(i)).
Previous reconstruction-based detectors generally distinguish OOD samples by comparing pixel-level quality ``degradation'' of the reconstruction for given input~\cite{taylor2018improving,schlegl2017unsupervised}.
However, without prior-knowledge about OOD samples, there is no guarantee for such quality degradation.
In contrast, \sood tries to synthesize In-D images instead of reconstructing the inputs, which is in line with the objective of the generative model.

The framework of \sood (Fig.~\ref{fig:related}(ii)) 
is inspired by~\cite{contraNet}, which detects adversarial examples (AE) by generating synthesized images conditioned on the output of the misled classifier.
AE detection is quite different from OOD detection because adversarial examples are In-D samples with imperceptible perturbations. The classification label itself is sufficient to train the generative model to differentiate AEs and benign samples. This is not the case for OOD samples, 
which motivates the proposed \sood solution for OOD detection, as detailed in Section~\ref{sec:overview}.


\subsection{OOD Detection with External OOD Data}
\label{sec:external_ood}
Recently, some researchers propose to involve data from other datasets to simulate OOD samples for model training.
Representative ``OOD-aware'' techniques include Outlier Exposure (OE)~\cite{hendrycks2018deep}, Maximum Classifier Discrepancy (MCD)~\cite{Yu_2019_ICCV}, and Unsupervised Dual Grouping (UDG)~\cite{yang2021semantically}. 
OE relies on large-scale purified OOD samples, whereas MCD and UDG only need extra unlabeled data, which contains both In-D and OOD data.
However, all of them are classification-based methods, where external data are used to train a modified classifier model.
Following the same unlabeled extra data setting of MCD and UDG, we present that including extra training data (In-D, OOD mixture) into the training process can further improve \sood's performance.
Since our \sood works independently with the original classifiers, \sood will not degrade the accuracy of the original classifiers.
We provide the detailed description in Section~\ref{sec:extra_ood} and experimental results in Section~\ref{sec:experiments}.

\subsection{Open Set Recognition}
A similar problem to OOD detection is the so-called Open Set Recognition (OSR) problem, 
which aims to distinguish the known and unknown classes\cite{oza2019c2ae,yang2021generalized}. 
Several existing works~\cite{oza2019c2ae,guo2021conditional,ge2017generative,neal2018open} targeted on OSR also employ generative models, whereas differ from \sood significantly. Specifically, OSRCI~\cite{neal2018open} uses Generative Adversarial Network (GAN) as data augmentation to train the classifier; C2AE~\cite{oza2019c2ae} and CVAECapOSR~\cite{guo2021conditional} are conditional VAE/Autoencoder-based detectors.   C2AE identifies outliers based on reconstruction errors, and requires K-time inference to give the final decision. In contrast, \sood infers once and makes the decision based on semantic contradiction. CVAECapOSR use Conditional VAE (CVAE) to model the distribution of In-D samples and detect outliers at latent space (i.e., without using the generator), whereas \sood detects at image space. 
\section{Proposed Method}
\label{sec:overview}

\subsection{Design Goals}
\label{sec:design_goal}
This work considers the scenario where we have an In-D data trained classifier ($\mathcal{C}$), which needs to be deployed in the wild. Consequently, the classifier will be threatened by OOD samples. 
We aim at building an OOD detection method, which can identify the OOD samples effectively without compromising the classification accuracy of the classifier.
Our model is assumed to have access to the predicted label, $y$, but we do not modify any part of the classifier, including but may not limit to the architecture and the trained weights.
As a result, our method can be a plug-and-play detector that easily cooperates with classifiers.

\begin{figure}[t]
	\centering\includegraphics[width=0.9\linewidth]{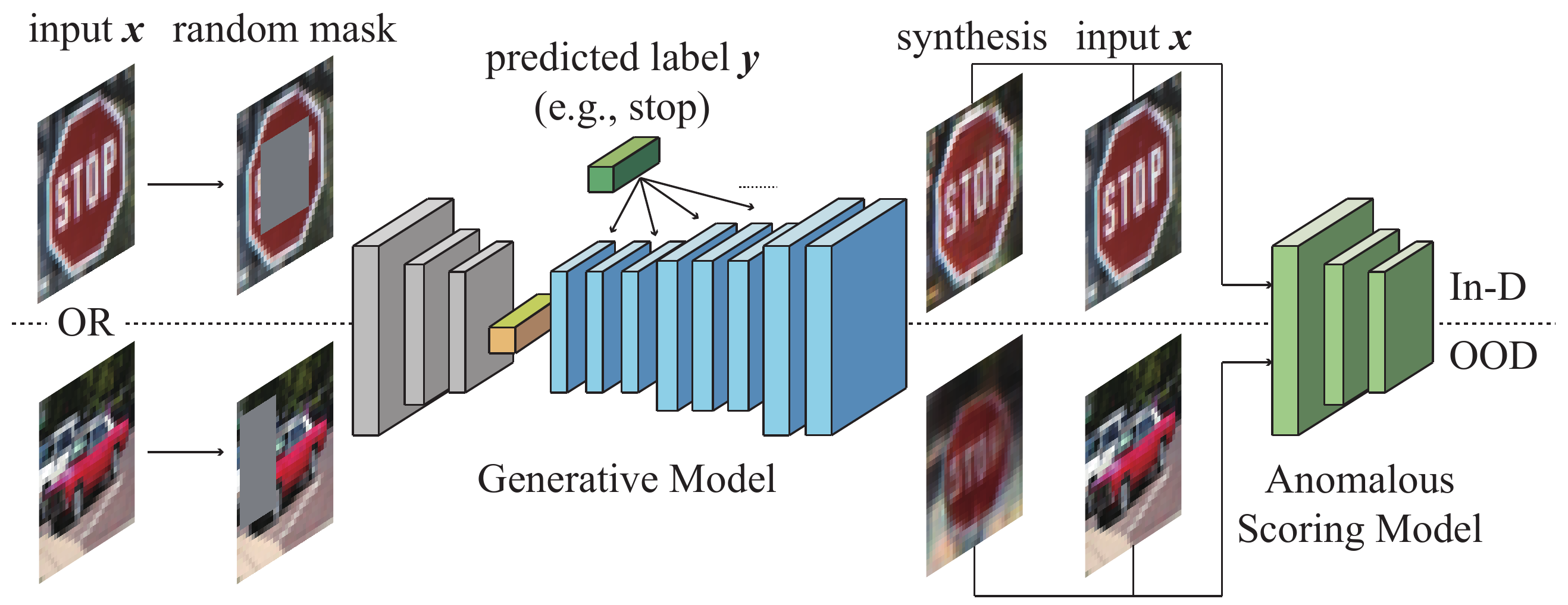}
	\caption{Pipeline of \sood. We first mask a portion of the input image. Next, a generative model synthesizes the masked image to a new image conditioned on  $y$, and then an anomalous scoring model measures the semantic difference between the input image and the synthesized one for OOD detection.}
	\label{fig:overview}

\end{figure}

\subsection{Method Overview}

As pointed out by \cite{yang2021generalized}, OOD samples ($x_o\in \mathcal{O}$) are defined by label-shifted samples or samples with non-overlapping labels w.r.t the training data, or In-Ds ($x_{in} \in \mathcal{I}$).
Hence, the semantics of any OOD sample  contradicts with any In-D sample.
This is the observation that motivates us to design a framework for OOD detection by spotlighting their semantic discrepancy.

Fig.~\ref{fig:overview} depicts the overview of our method.
The proposed Masked Out-of-Distribution Catcher (\sood) contains three stages: \textit{randomly masking, generative synthesis} and \textit{scoring}. 
Specifically, we first randomly mask the input image $x$ as $x_m=\mathbf{M}(x)$, where $\mathbf{M}(\cdot)$ indicates the randomly masking operation.
Then, we apply a generative model, \G, to synthesize a new image, $x'$, by taking $x_{m}$ as the template and conditioning on the label $y$.
Finally, we apply an anomalous scoring model to judge the discrepancy between the input and its synthesis.

Through masking, $x_{m}$ will partially lose its original semantic meaning, and thus leave more space for \G to synthesize new content.
With $y$ as the condition, the newly synthesized content should be consistent with the semantic meaning indicated by $y$.
Note that, we use the ground truth label $y$ to train the generative model and use the output from the classifier when inference, i.e., $y=\mathcal{C}(x)$.

Here, we analyze different situations with In-D or OOD samples.
On the one hand, if an In-D sample $x_{in}$ comes, the predicted label $y$ matches $x_{in}$'s intrinsic semantic meaning appropriately. 
Although the input image, $x_{in\cdot m}$, is partially masked, there should be some visual clue related to its semantic meaning, e.g., wings of a bird or paws of a dog.
As a result, \G can synthesize $x'_{in}$ quite faithful to $x_{in}$.
As exemplified in the upper half of Fig.~\ref{fig:overview}, the synthesis of the ``stop'' sign can be very close to the original input.
On the other hand, when it comes to an OOD sample $x_{o}$, the predicted label provided by the classifier is irreverent to $x_o$'s semantic meaning. 
Even with $x_{o}$ as a template, the generative model will try its best to synthesize contents related to the semantic label.
As a result, the mismatch of semantic meaning between input and label can be spotlighted by the discrepancy between input and its synthesis.
As the example shows in the lower part in Fig.~\ref{fig:overview}, if an OOD sample (\textit{car}) is wrongly predicted as a ``stop'' sign. 
The synthesis will be highly related to the ``stop'' sign rather than the original image.

Through such conditional synthesis, we can spotlight the discrepancy caused by OOD samples.
Thus OODs can be easily distinguished by comparing the pair of input with its synthesis, $(x, x')$.


\subsection{Masking Mechanism}
In \sood, the generative model uses the input image as a template and synthesizes an image with the same semantic meaning as the given label.
A high-quality synthesis can better highlight the contradiction.
However, due to the intrinsic contradiction between the input image and the label for OOD, too much information from the input image can degrade the quality of generation.
Therefore, we propose to apply masking on $x$ to remove some redundant information while leaving more space for the generative model to synthesize.

The use of masking follows the key motivation of \sood in OOD detection, which applies generation for synthesis rather than reconstruction.
Previous reconstruction-based methods (e.g., \cite{taylor2018improving,schlegl2017unsupervised}) tend to reconstruct the image based on pixel-level dependency.
In practice, the assumption that an OOD sample cannot be reconstructed well may not hold since they do not consider any semantics.
However, our generative model aims at semantic synthesis.
The masking mechanism can cooperate with the predicted label from the classifier to spotlight the contradiction caused by OOD.

The contribution of randomly masking is twofold:
\textbf{1)} masking the input image can encourage the generative model to better depict the semantic meaning of the given label on the synthesis, especially to an OOD sample;
\textbf{2)} masking, as a typical data augmentation method, can encourage the encoder to summarize the features of the input from a holistic perspective, thus improve the quality of synthesis, especially when synthesizing with In-Ds as templates.
Obviously, the above two aspects both contribute to apart the behavior of In-D and OOD.
As a result, a large discrepancy lies in the OOD sample and its synthesis.




\subsection{Generative Model}
The Generative model is responsible for generating a synthesis by taking both the masked input $x_m$ and the pre-assigned semantic label $y$ into consideration.
As shown in Fig.~\ref{fig:overview}, we select the Encoder ($\mathbf{E}$) and Decoder ($\mathbf{D}$) architecture as the generative model, i.e. $\mathbf{G} = \mathbf{E} \cdot \mathbf{D}$.
This architecture is inspired by~\cite{contraNet}. 
The encoder \E acts as a feature extractor (as shown by the gray part in Fig.~\ref{fig:overview}).
By taking  the masked image $x_m$ as input, \E is expected to capture necessary low-level features and encoder them as a latent vector, $z=\mathbf{E}(x_m)$.
As done by VAE~\cite{kingma2013auto}, we use the KL Divergence to regulate the latent vector $z$, which can be formulated as \equ{\ref{eq:KLD}}.
\begin{equation}
	\mathcal{L}_{KLD} = D_{KL}[\mathcal{N}(\mu(x_m),\varSigma (x_m))\| \mathcal{N}(0,1)]\text{,}\label{eq:KLD}
\end{equation}
where $\mathcal{N}(\mu, \varSigma)$ indicates the Gaussian distribution with respect to $\mu$ and $\mathcal{\varSigma}$. We use the reparameterization trick from VAE on the latent variable $z$ during training, $z=\mu(x_m)+\varSigma({x_m})\cdot\epsilon$, where $\epsilon\thicksim \mathcal{N}(0, 1)$.

The decoder, \D, is trained to generate a synthesis $x'=\mathbf{D}(z, y)$.
The given semantic label $y$ is used to control the semantic meaning of the synthesis, while $z$ is used to provide low-level features from the template image $x$.
This synthetic target is fulfilled through the class-conditional batch normalization layer~\cite{de2017modulating}.
This layer is usually used in conditional image generation~\cite{miyato2018cgans,zhang2019self}.
Since the normalization is determined by the given semantic label $y$, the semantic meaning of the synthesis can be highly dependent on it.
As a result, if the semantic meaning of $x$ is consistent with $y$ (in the case of In-D samples), the synthesis can be highly close to $x$.
However, if input an OOD sample, the semantic contradiction between the input image and the label will lead the synthesis to be far away from the input image, thus spotlighting the contradiction.

We implement \D based on the generator architecture proposed in~\cite{brock2018large}.
We apply the classic $\ell_{1}$, $\ell_{2}$ and $\mathcal{SSIM}$~\cite{sara2019image} as part of loss items to constrain that $x'$ resembles $x$. 
Furthermore, we adopt the U-net based discriminator~\cite{schonfeld2020u} to operate an adversarial loss on the training process to further improve the quality of the synthetic image. 
Compared with the vanilla discriminator, this U-net based one can additionally provide a per-pixel real/fake map to locate the fake parts in the image.
Therefore, the generative model can be trained to focus on both local and global features with more realistic details.
Due to space limitations, we detailed the training process and corresponding objective functions of \G in Appendix. 

\subsection{Anomalous Scoring Model}
As analyzed in the former sections, \sood can generate high-quality syntheses in terms of similarity for In-D samples, but not for OOD samples. 
To distinguish OODs from In-Ds, we develop an anomalous scoring model.
The proposed anomalous scoring model is built on two types of scorers: one is the \emph{conditional binary classifier}, and the other is \emph{Image Quality Assessment models} (IQA)~\cite{zhang2018unreasonable}.
Both 
can provide assessments of the syntheses by referring to 
input images.

\subsubsection{\textit{Conditional Binary Classifier.}} 
\label{sec:binary_classifier}
Identifying the semantic mismatch lies between OOD and its synthesis can be seen as a binary classification task.
With the trained generative model, we can train a binary classifier for this in a supervised way.
This is feasible because the binary classifier can learn to identify OODs by the similarity between the given image and its conditional synthesis.
We also provide the semantic label to the classifier for judgement, this can further ease the distinguishing procedure.

Note that, 
we do not rely on OOD data through training.
To mitigate the lack of OOD samples during training, we leverage the In-D samples with the synthetic results under mismatched labels to simulate the behavior of OOD samples. 
As a result, the binary classifier can learn to identify the semantic mismatch, which is the spotlighted feature for OOD samples.
Moreover, we condition the binary classifier on the semantic label via the projection layer~\cite{miyato2018cgans}. 
With the prior knowledge of the class, the binary classifier can learn a fine-grained decision boundary for each class, leading to better performance. 

Another good property of this training strategy is that there is no need for specific information from the DNN model, here the classifier ($\mathcal{C}$),  to be protected.
Actually, the judgement of the binary classifier is only conditioned on given label, which is independent of the DNN model. 
Therefore, the trained conditional binary classifier can fit various DNN models in a plug-and-play manner.

\setlength{\intextsep}{5pt}%
\begin{wrapfigure}[14]{r}{0.5\textwidth}
	
		\includegraphics[width=0.5\textwidth]{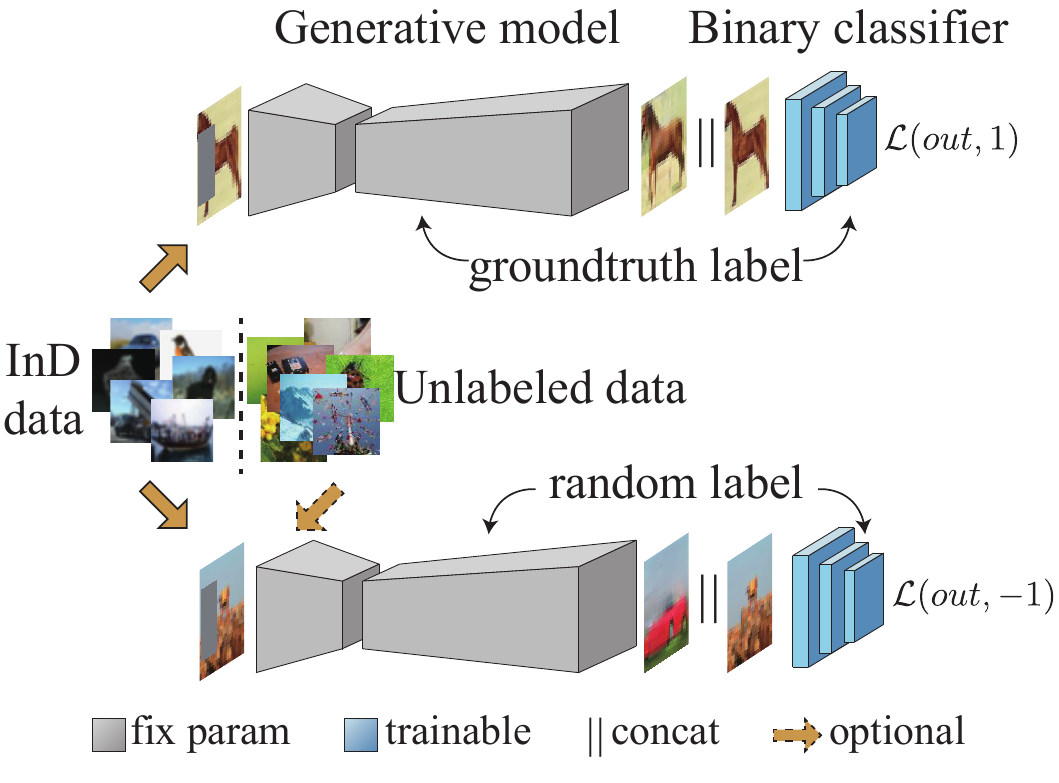}
		
		\caption{Training pipeline of the proposed Conditional Binary Classifier.}
		\label{fig:binary_classifier}
\end{wrapfigure}

Fig.~\ref{fig:binary_classifier} demonstrates the training process of the proposed conditional binary classifier, $\mathcal{C}_b$. 
\emph{hinge loss} serves as the objective, as formulated in \equ{\ref{eq:hinge_loss}}, where the $x_y'$ indicates the synthesis generated under the groundtruth label $y$ of $x$, and the $(x, x_y')$ is used as the positive pair.
$x_{y'}'$ depicts the synthesis generated under a randomly sampled mismatched semantic label $y'\neq y$ , then $(x, x_{y'}')$ is used as the negative pair during training. 
During inference, $\mathcal{C}_{b}$ is used to score the input image pair.
The score can be used to flag OOD samples by a given threshold.
\begin{equation}
		\label{eq:hinge_loss}
		\mathcal{L}_{\mathcal{C}_{b}} = ReLU(1-\mathcal{C}_{b}((x,x_{y}'), y))+ ReLU(1+\mathcal{C}_{b}((x, x_{y'}'), y'))
\end{equation}





\subsubsection{\textit{Image Quality Assessment Models.}}
\label{sec:iqa}
IQA models~\cite{zhang2018unreasonable} are widely adopted to evaluate the perceptual quality of a synthesis by referring to the source image in many computer vision tasks, such as denoising, super-resolution and compression.
Here, we apply IQA models as the perceptual metric for the quality of synthesis, forming part of our anomalous scoring model.
Since our generative model has already highlighted the contradiction caused by OOD through conditional synthesis, IQA models can be directly applied for detection.

In all, the \textit{ Conditional Binary Classifier} and \textit{ IQA} models work in a cascade way, where any scorer flags an OOD can lead to the final rejection. 
Different scoring mechanisms can evaluate the quality of generation from different perspectives, thus supporting each other for better performance than anyone alone. 

\subsection{Learning with Unlabeled Data}
\label{sec:extra_ood}
Recently, researchers have shown that including OOD data into the training process can improve the performance of OOD detection~\cite{hendrycks2018deep,Yu_2019_ICCV}.
However, they may rely on well labeled data that need to manually identify In-D and OOD elements~\footnote{Note that, images from another dataset are not necessarily to be OOD w.r.t semantic meaning~\cite{yang2021semantically}.}, e.g., ~\cite{hendrycks2018deep}.
On the contrary, unlabeled data can be easily collected from various sources with little cost.
Actually, \sood can be further improved with unlabeled data.

The Conditional Binary Classifier introduced in Section~\ref{sec:binary_classifier} is trained by treating In-D data with synthesis of mismatched semantic labels as negative pairs.
This procedure can be directly combined with external data.
Note that, our negative pair only require that the semantic label used for synthesis is different from that of input image.
To eliminate possible In-D samples from unlabeled data, we can apply a classifier trained with In-D data to generate pseudo label for unlabeled data.
Then, during training, when we randomly sample the mismatched semantic label, we uniformly sample from all possible labels other than the pseudo label.
In this way, the sampled semantic label is ensured to be mismatched for both OOD and some possible In-D data from the unlabeled data.
Therefore, all of them can be utilized correctly to improve the performance of \sood.
\section{Experiments}
\label{sec:experiments}

\subsection{Evaluation Settings}
\textbf{\textit{Benchmarks.}}  We evaluate \sood on the most recent semantic OOD detection benchmarks, SC-OOD benchmarks~\cite{yang2021semantically}. 
SC-OOD benchmarks provide extensive semantic-level OOD detection settings for evaluation. 
Specifically, from SC-OOD, images from different datasets are filtered to ensure that only those containing different semantic meanings are considered as OOD samples. 
SC-OOD focus on the semantic difference between samples, thus being more practical for the real-world model deployment than other previous OOD benchmarks\cite{hendrycks2016baseline,liang2018enhancing,hendrycks2018deep}, which are built by setting one dataset as In-D and all others as OOD.

\noindent\textbf{\textit{Datasets.}}
Following the settings in \cite{yang2021semantically}, we employ \cifar~\cite{krizhevsky2009learning}, \cifarh~\cite{krizhevsky2009learning} as In-D samples, respectively, and others as OOD samples.
When setting \cifar as In-D, we employ six datasets as OOD datasets, including \svhn\cite{37648}, \cifarh\cite{krizhevsky2009learning}, \texture~\cite{cimpoi14describing}, \places~\cite{zhou2017places}, \lsun~\cite{yu2016lsun} and \tinyimagenet~\cite{tiny}.
For \cifarh benchmarks, the OOD datasets are the same as that of \cifar, except for swapping \cifarh for \cifar as OOD.     

\subsection{Evaluation Metrics}
We employ FPR@TPR95\%, AUROC, AUPR, and Classification Accuracy as the evaluation metrics following \cite{yang2021semantically,EBO}. 
In this paper, unless otherwise specified, we denote the In-D as Positive (P), and the OOD as Negative (N).

\noindent\textit{\textbf{FPR@TPR95\%}} presents the False Positive Rate (FPR) when the True Positive Rate (TPR) equals 95\%. 
This metric reflects the ratio of falsely identified OOD when most of In-D samples are correctly recognized.

\noindent \textbf{\textit{AUROC}}.
The Area Under Receiver Operating Characteristic curve (AUROC) is an overall evaluation metric to reflect the detection capability of a detector.

\noindent \textbf{\textit{AUPR-In}} AUPR calculates the Area Under the Precision-Recall curve. AUPR is a complementary metric that reflects the impact of imbalanced datasets.
For AUPR-In metric, In-D samples are denoted as positive samples. 

\noindent \textbf{\textit{AUPR-Out}} indicates the same measure as AUPR-In mentioned above, whereas the OOD samples are deemed as positive during calculating AUPR-Out.

\noindent \textbf{\textit{Classification Accuracy}} presents the classifier's performance on the In-D samples. It indicates the impact on the original classifier caused by OOD detector.

\subsection{Experimental Results}
\label{sec:table_results}
We evaluate \sood with two settings: \textbf{1)} \sood trained with In-D dataset only; \textbf{2)} \sood trained with external unlabeled data (Section~\ref{sec:extra_ood}).
We report the results in Table~\ref{tab:cifar10} and Table~\ref{tab:cifar100}, respectively. 
Experiments are performed with ResNet18~\cite{he2016deep} classifier\footnote{For more results under other classifier architectures (WRN28~\cite{zagoruyko2016wide}, DenseNet~\cite{8721151}), please refer to Appendix.} for fair comparison.


\noindent\textbf{\textit{Main Results.}} As shown in Table~\ref{tab:cifar10} and  Table~\ref{tab:cifar100}, experimental results indicate that \sood outperforms or at least on par with SOTA methods on \cifar/\cifarh benchmarks without/with external training data.
Since our method detects OODs relying on their semantic-level mismatching instead of low-level distribution shift, the performance of \sood is stable across various OODs. 
As a plug-and-play model, \sood causes no  classifier performance degradation. 

\begin{table}[tb]
    \caption{OOD Detection Performance on \cifar as In-D without using external OOD data for training. All the values are in percentages. $\uparrow$/$\downarrow$ indicates higher/lower value is better. The best results are in \textbf{bold}. We also add our results with external data in {\color{gray}gray}.} 
	\label{tab:cifar10}
	\fontsize{6}{9}\selectfont
	\begin{tabularx}{\textwidth}{ P{1.5cm} P{2cm} s s s s P{1.7cm} }
	\hline 
	
	\begin{tabular}{@{}c@{}}\bf{Detection} \\ \bf{Methods} \end{tabular}
	& \begin{tabular}{@{}c@{}}\bf{OOD} \end{tabular}
	& \begin{tabular}{@{}c@{}} \bf{FPR@} \\ \bf{TPR95\%} \bf{$\downarrow$} \end{tabular} 
	& \begin{tabular}{@{}c@{}} \bf{AUROC}   \bf{$\uparrow$} \end{tabular}
	& \begin{tabular}{@{}c@{}} \bf{AUPR} \\ \bf{In}  $\uparrow$ \end{tabular}
	&  \begin{tabular}{@{}c@{}} \bf{AUPR} \\ \bf{Out} $\uparrow$ \end{tabular}
	&  \begin{tabular}{@{}c@{}} \bf{Classification} \\ \bf{Accuracy}  $\uparrow$ \end{tabular}\\ 
    \hline

	\multirow{7}{*}{{\bf{ODIN}}\cite{liang2018enhancing}}
	& \svhn &52.27 &83.26 &63.76 &92.60 &\textbf{95.02}  \\
	& \cifarh  &61.19 &78.40 & 73.21 & 80.99 &\textbf{95.02}  \\
	& \tinyimagenet &59.09 &79.69 &79.34 &77.52 &\textbf{92.54} \\
	& \texture &42.52 &84.06 &86.01 &80.73 &\textbf{95.02} \\
	& \lsun &47.85 &84.56 &81.56 &85.58 &\textbf{95.02}\\
	& \places &53.94 &82.01 &54.92 &93.30 &\textbf{93.87} \\\cline{2-7} 
	&\textbf{Mean/Std} &52.00&82.00/2.48&73.13/11.79&85.12/6.59&\textbf{ 94.42/1.03} \\
	 
	\hline
	\multirow{7}{*}{{\bf{EBO}}\cite{EBO}}
	& \svhn  &\textbf{30.56} &92.08 &80.95 &96.28 &\textbf{95.02}\\
	& \cifarh &56.98 &79.65 &75.09 &81.23 &\textbf{95.02} \\
	& \tinyimagenet &57.81 &81.65 &81.80 &78.75 &\textbf{92.54} \\
	& \texture &52.11 &80.70 &83.34 &75.20 &\textbf{95.02} \\
	& \lsun &50.56 &85.04 &82.80 &85.29 &\textbf{95.02} \\
	& \places &52.16 &83.86 &58.96 &93.90 &\textbf{93.87} \\\cline{2-7}
	& \textbf{Mean/Std} &50.03&83.83/4.51&77.16/9.40&85.11/8.44&\textbf{ 94.42/1.03} \\

	\hline
	\multirow{7}{*}{\bf{Ours}}
	& \svhn & 37.72/{\color{gray}24.27} &\textbf{92.99}/{\color{gray}95.93} &\textbf{87.43}/{\color{gray}92.98} &\textbf{96.70}/{\color{gray}98.05} &\textbf{95.02} \\
	& \cifarh &\textbf{42.32}/{\color{gray}39.92} &\textbf{89.88}/{\color{gray}91.45} &\textbf{89.75}/{\color{gray}91.54} &\textbf{90.24}/{\color{gray}91.73} &\textbf{95.02} \\
	& \tinyimagenet &\textbf{40.60}/{\color{gray}32.41} &\textbf{90.57}/{\color{gray}93.34} &\textbf{90.59}/{\color{gray}93.63} &\textbf{90.76}/{\color{gray}93.41} &\textbf{92.54} \\
	& \texture &\textbf{26.12}/{\color{gray}6.86} &\textbf{94.15}/{\color{gray}98.69} &\textbf{96.33}/{\color{gray}99.29} &\textbf{91.68}/{\color{gray}97.71} &\textbf{95.02} \\
	& \lsun &\textbf{43.86}/{\color{gray}33.31} &\textbf{90.61}/{\color{gray}93.40} &\textbf{91.07}/{\color{gray}93.85} &\textbf{90.02}/{\color{gray}93.22} & \textbf{95.02}\\
	& \places & \textbf{42.34}/{\color{gray}35.51} &\textbf{90.16}/{\color{gray}92.77} &\textbf{75.28}/{\color{gray}82.25} &\textbf{96.83}/{\color{gray}94.82} & \textbf{93.87}\\\cline{2-7}
	& \textbf{Mean}&\textbf{38.83}/{\color{gray}28.71} &\textbf{91.39}/{\color{gray}94.27} & \textbf{88.40}/{\color{gray}92.26} &\textbf{92.71}/{\color{gray}94.82} &\textbf{ 94.42} \\	
	& \textbf{Std} & - &\textbf{1.75}/\textcolor{gray}{2.61}&\textbf{7.07}/\textcolor{gray}{5.57}&\textbf{3.20}/\textcolor{gray}{2.56}&\textbf{1.03} \\

	\hline 
	\end{tabularx}

\end{table}

\begin{table}[tb]
    \centering
    \caption{OOD Detection Performance on \cifarh as In-D with \tinyimagenet as external data for training. All the values are in percentages. $\uparrow$/$\downarrow$ indicates higher/lower value is better. The best results are in \textbf{bold}. We also add our results without external data in {\color{gray}gray}.}
    \label{tab:cifar100}

    \fontsize{6}{9}\selectfont
	\begin{tabularx}{\textwidth}{ P{1.5cm} P{2cm} s s s s P{1.7cm} }
	\hline 
	
	\begin{tabular}{@{}c@{}} \bf{Detection} \\ \bf{Methods} \end{tabular}
	& \begin{tabular}{@{}c@{}} \bf{OOD} \end{tabular}
	& \begin{tabular}{@{}c@{}} \bf{FPR@} \\ \bf{TPR95\%}  \bf{$\downarrow$} \end{tabular} 
	& \begin{tabular}{@{}c@{}} \bf{AUROC}  \bf{$\uparrow$} \end{tabular}
	& \begin{tabular}{@{}c@{}} \bf{AUPR} \\ \bf{In}  $\uparrow$ \end{tabular}
	&  \begin{tabular}{@{}c@{}} \bf{AUPR} \\ \bf{Out}  $\uparrow$ \end{tabular}
	&  \begin{tabular}{@{}c@{}} \bf{Classification} \\ \bf{Accuracy}  $\uparrow$ \end{tabular}\\

	 \hline

	\multirow{7}{*}{{\bf{MCD}}\cite{Yu_2019_ICCV}}
	& \svhn &85.82 &76.61 &65.50 &85.52 &68.80 \\
	& \cifar &87.74 &73.15 &76.51 &67.24 &68.80 \\
	& \tinyimagenet &84.46 &75.32 &85.11 &59.49 &62.21 \\
	& \texture &83.97 &73.46 &83.11 &56.79 &68.80 \\
	& \lsun &86.08 &74.05 &84.21 &58.62 &67.51 \\
	& \places &82.74 &76.30 &61.15 &87.19 &70.47 \\\cline{2-7}
	& \textbf{Mean/Std} & 85.14 &74.82/1.47&75.93/10.31&69.14/13.81&67.77/2.88\\
	\hline
	\multirow{7}{*}{{\bf{OE}}~\cite{hendrycks2018deep}}
	 & \svhn &68.87 &84.23 &75.11 &91.41 &70.49 \\
	 & \cifar &79.72 &78.92 &81.95 &74.28 &70.49 \\
	 & \tinyimagenet &83.41 &76.99 &86.36 &60.56 &63.69 \\
	 & \texture &86.56 &73.89 &84.48 &54.84 &70.49 \\
	 & \lsun &83.53 &77.10 &86.28 &60.97 &69.89 \\
	 & \places &78.24 &79.62 &67.13 &88.89 &72.02 \\\cline{2-7}
	 & \textbf{Mean/Std}&80.06&78.46/3.46&80.22/\textbf{7.66}&71.83/15.58&69.51/2.94\\
	 \hline
	
    \multirow{7}{*}{{\bf{UDG}}~\cite{yang2021semantically}}
	& \svhn &60.00 &88.25 &\textbf{81.46} &93.63 &68.51 \\
	& \cifar &83.35 &76.18 &78.92 &71.15 &68.51 \\
	& \tinyimagenet &81.73 &77.18 &86.00 &61.67 &61.80 \\
	& \texture &75.04 &79.53 &87.63 &65.49 &68.51 \\
	& \lsun &78.70 &76.79 &84.74 &63.05 &67.10 \\
	& \places &73.89 &79.87 &65.36 &89.60 &69.83 \\\cline{2-7}
	& \textbf{Mean/Std}&75.45&79.63/4.48&80.69/8.14&74.10/14.01&67.38/\textbf{2.87}\\
	 
	\hline
	\multirow{7}{*}{\bf{Ours}}
	& \svhn & \textcolor{gray}{58.16}/\textbf{51.60} &\textcolor{gray}{87.38}/\textbf{88.99} &\textcolor{gray}{78.25}/80.89 & \textcolor{gray}{93.81}/\textbf{94.81}& \textbf{76.65}\\
	& \cifar &\textcolor{gray}{54.31}/\textbf{50.17} &\textcolor{gray}{85.91}/\textbf{87.76} &\textcolor{gray}{86.27}/\textbf{88.18} &\textcolor{gray}{85.91}/\textbf{87.79} & \textbf{76.65}\\
	& \tinyimagenet & \textcolor{gray}{55.33}/\textbf{46.07}&\textcolor{gray}{86.95}/\textbf{89.42} &\textcolor{gray}{87.55}/\textbf{89.73} &\textcolor{gray}{86.67}/\textbf{89.28} & \textbf{69.56}\\
	& \texture & \textcolor{gray}{46.70}/\textbf{42.22}&\textcolor{gray}{89.20}/\textbf{90.56} &\textcolor{gray}{93.48}/\textbf{94.43} &\textcolor{gray}{83.28}/\textbf{85.13} & \textbf{76.65}\\
	& \lsun & \textcolor{gray}{53.43}/\textbf{47.85}&\textcolor{gray}{87.98}/\textbf{89.96} & \textcolor{gray}{88.82}/\textbf{90.33}&\textcolor{gray}{87.32}/\textbf{89.23} & \textbf{76.10}\\
	& \places &\textcolor{gray}{54.20}/\textbf{47.72}&\textcolor{gray}{87.41}/\textbf{89.30} &\textcolor{gray}{71.68}/\textbf{74.83} &\textcolor{gray}{95.78}/\textbf{96.48} & \textbf{77.56}\\\cline{2-7}
	& \textbf{Mean}& \textcolor{gray}{53.69}/\textbf{47.61} &\textcolor{gray}{87.47}/\textbf{89.33} & \textcolor{gray}{84.34}/\textbf{86.40} & \textcolor{gray}{88.80}/\textbf{90.45} &\textbf{75.53} \\
	& \textbf{Std} & - &\textcolor{gray}{0.95}/\textbf{1.09}&\textcolor{gray}{7.19}/7.94&\textcolor{gray}{4.33}/\textbf{4.89}&2.96 \\
	\hline
	\end{tabularx}

\end{table}

\noindent\textbf{\textit{\cifar Benchmark.}}  
Table~\ref{tab:cifar10} reports the detection performance of ODIN~\cite{liang2018enhancing}, EBO~\cite{EBO} and \sood on \cifar benchmarks. 
ODIN and EBO are developed without external OOD data.
When comparing these two methods, we implemented \sood in the same setting.
We additionally report results with external data for training to show the improvements.
As shown in Table~\ref{tab:cifar10}, \sood outperforms ODIN and EBO in most cases. 
For example, for the AUROC, which reflects the overall performance of a detector, \sood outperforms baselines on all six OOD benchmarks.
Furthermore, \sood presents a more stable performance across all OOD datasets.
As for statistics, we report the standard deviation (Std) for each metric~\footnote{FPR@TPR95\% is only a single point on the PR curve. It may not reflect the overall performance in terms of standard deviation.}.
As in Table~\ref{tab:cifar10}, the Std of \sood can be much lower than baselines.
To be more specific, EBO performs quite well when encounter OODs sourcing from \svhn (AUROC=92.08\%), but when it comes to \cifarh, the AUROC drops by 12\% to 79.65\%. 
On the contrary, the AUROC for \sood are all around a high mean value.
This stability may due to that \sood utilize semantic information, which is exactly the definition of OOD.
Comparing to the classification features or other low-level features, the semantic contradiction exists more generally.

As discussed in Section~\ref{sec:extra_ood}, \sood can be equipped with external OOD data for better detection ability.
In Table~\ref{tab:cifar10}, we report the performance for \sood using \tinyimagenet as external unlabeled training data in \textcolor{gray}{gray}.
These results evidence the effectiveness of the training strategy with external data for \sood, where significant performance improvements can be found.
Due to space limitation, for this setting, we only compare results with baselines on \cifarh (see below) and leave that on \cifar benchmarks in  Appendix.


\noindent\textbf{\textit{\cifarh Benchmark.}}
OE~\cite{hendrycks2018deep}, MCD~\cite{Yu_2019_ICCV} and UDG~\cite{yang2021semantically} rely on \tinyimagenet as external OOD data for training. 
We use them as baselines to show the effectiveness of \sood under this setting.
Note that \sood treats all external data as unlabeled during training.

Table~\ref{tab:cifar100} shows the results on \cifarh benchmarks.
As can be seen, though taking advantage of external data, MCD, OE and UDG suffer from the complicated \cifarh dataset.
By contrast, \sood's performance on \cifarh is comparable with that on \cifar against same OOD shown in Table~\ref{tab:cifar10}.
This is due to \sood's detection mechanism, which relying on the semantic mismatch.
Even when the In-D dataset becomes complicated, the semantic mismatch in OODs is still obvious.
For methods detect OODs based on DNN-extracted features, e.g. MCD, OE, UDG, they may suffer from poor decision boundaries as the number of classes increases.
Similarly as in the case on \cifar, to show the effectiveness of the proposed training strategy with external data, we also report the performance for \sood without external data in \textcolor{gray}{gray}.
The improvement can be seen on all metrics across all OOD settings.
For space limitation, we leave the comparison on \cifarh benchmarks without external data in Appendix.

\subsection{Comparison with Open Set Recognition Methods}
The experimental protocol in OSR is to randomly selecting $K$ classes from a specific $n$-class dataset  as ``known'' classes and the left $n-K$ as ``unknown'' classes. To make a fair comparison, we retrain \sood under the experimental settings in CVAECapOSR~\cite{guo2021conditional}, where only 4 or 6 classes from \cifar are used for training.
We report AUROC scores in Table~\ref{tab:osr}, and 
the results for other methods are from CVAECapOSR~\cite{guo2021conditional}. As can be seen,  \sood outperforms these methods in all three settings, especially in the original \osrcifar setting, where \sood outperforms the second best method about 6\%. 

\begin{table}[tbp]
	\centering
	\caption{Comparison with Open Set Recognition methods. 
	AUROC scores on the detection of known and unknown classes.
	\osrcifar indicates splitting \cifar to 6 known classes, and 4 unknown. \osrcifar+$N$ samples known 4 classes form \cifar, $N$ unknown classes from \cifarh. The \textbf{bold} indicates the best. For more details about dataset splits, please refer to CVAECapOSR~\cite{guo2021conditional}.
	}
	\label{tab:osr}
	\setlength{\tabcolsep}{6mm}{
	\begin{tabular}{@{}cccc@{}}
	\hline

Method     & \osrcifar & \osrcifar+10 & \osrcifar+50 \\
	\hline
OSRCI~\cite{neal2018open}      & 69.9$\scriptstyle{\pm 3.8}$     & 83.8     & 82.7     \\
	C2AE~\cite{oza2019c2ae}       & 71.1$\scriptstyle{\pm 0.8}$    & 81.0$\scriptstyle{\pm 0.5}$     & 80.3$\scriptstyle{\pm 0.0}$     \\
	CVAECapOSR~\cite{guo2021conditional} & 83.5$\scriptstyle{\pm 2.3 }$  & 88.8$\scriptstyle{\pm 1.9}$ & 88.9$\scriptstyle{\pm 1.7}$  \\
	\hline
	\textbf{\sood (ours)}       &  \textbf{89.48}$\scriptstyle{\pm 0.50}$       & \textbf{89.36}$\scriptstyle{\pm 0.74}$ & \textbf{89.23}$\scriptstyle{\pm 0.19}$    \\
	\hline
	\end{tabular}
	
	}
	\end{table}


\begin{table}[tbp]
	\centering
    \caption{OOD Detection Performance under different combinations of anomalous scorers. \sood is trained on \cifar(In-D) without external data. OODs are from \cifarh.  All the values are in percentages. $\uparrow$/$\downarrow$ indicates the higher/lower value is better. The best results are in \textbf{bold}.}
	\label{tab:scorer}
	\fontsize{8}{9}\selectfont
	\begin{tabularx}{\textwidth}{ p{3.1cm} s s s s   }
	\hline 
	
	\begin{tabular}{@{}c@{}}\fontsize{8}{9} Anomalous Scorer \end{tabular}
	& \begin{tabular}{@{}c@{}}  FPR@TPR95\%\bf{$\downarrow$} \end{tabular} 
	& \begin{tabular}{@{}c@{}} \fontsize{8}{9}AUROC\bf{$\uparrow$} \end{tabular}
	& \begin{tabular}{@{}c@{}} AUPR-In $\uparrow$ \end{tabular}
	&  \begin{tabular}{@{}c@{}} AUPR-Out$\uparrow$ \end{tabular}\\
    \hline
		$\mathcal{C}_b$							& 42.80 & 89.13 & 88.58 & 89.85 \\
		$LPIPS$               					& 76.62 & 73.93 & 72.68 & 73.23 \\
		$DISTS$                  				& 82.03 & 72.14 & 71.83 & 70.35 \\
		$\mathcal{C}_b+LPIPS$     				& 42.14 & 89.49 & 89.18 & 90.11 \\
		$\mathcal{C}_b+DISTS$     				& 42.31 & 89.35 & 89.09 & 89.98 \\
		$LPIPS+DISTS$       					& 76.06 & 74.78 & 74.25 & 73.86 \\
		\textbf{$\mathcal{C}_b+LPIPS+DISTS$}	& \textbf{41.95} & \textbf{89.57} & \textbf{89.30} & \textbf{90.16} \\
	\hline
	\end{tabularx}

\end{table}

\subsection{Ablation Study}
\label{sec:ablation}
In this section, we first analyze the effectiveness of every anomalous scorer and how scorers cooperate to achieve the final decent performance. Then we conduct experiments on
masking, and give insights on masking's effectiveness.

\noindent\textbf{\textit{Anomalous scoring model.}}
\label{sec:ablation_score}
Table~\ref{tab:scorer} summarizes the performances of every scorer, every two scorers and all three scorers working together. 
As can be seen in Table~\ref{tab:scorer}, the $\mathcal{C}_b+LPIPS+DISTS$ combination wins the best detection performance in terms of all evaluation metrics, which means our proposed $\mathcal{C}_b$ has a high flexibility to cooperate with IQA models (see Section~\ref{sec:iqa}). 
We can also observe that coupling scorers usually lead to a better detection capability than that of any single scorer within the coupling. 
However, adding extra scorers inevitably increases computational and memory overheads. 
The cost of basic version of \sood, i.e. $\mathcal{C}_{b}$, \E, \D, is relatively small, i.e. Params 4.552M, MACs 0.408G, when compared to that of the widely adopted classifier architectures, e.g., ResNet50 (Params 23.251M, MACs 1.305G).
Due to space limitations, we detail the computational and memory costs of \sood and those of the baselines in the Appendix. 
In practice, \sood can achieve appropriate detection ability, by tailoring scorers in \sood's anomalous scoring model according to the application scenario, i.e. trade-off between the performance and costs, and we discussed this part in Appendix.

\noindent\textbf{\textit{Masking.}}
\label{ablation_masking}
To evidence the effectiveness of masking holistically, we conduct another ablation study without masking in training. As shown in Table~\ref{tab:masking}, the masking scheme  indeed plays an important role in the performance of \sood. Due to space limitation, we leave the ablation study on masking style in Appendix.

\begin{table}[tpb]
	\caption{Ablation study on Masking.  We set the masking ratio as 0.3 for ``Fixed High Ratio'' and ``Patched'', 0.1 for ``Fixed Low Ratio'', and that of ``Randomly'' varies from 0.1 to 0.3. \sood employs the \textbf{Randomly} masking style.}
	\label{tab:masking}
			\centering
			\renewcommand\arraystretch{0.9} 
			\setlength{\tabcolsep}{1mm}{
			\begin{tabular}{cccccc}
			\hline
					Training & Inference & FPR@95 ↓ & AUROC ↑ & AUPR In ↑ & AUPR Out ↑ \\ \hline
					\multirow{2}{*}{\rotatebox{90}{\makecell[c]{w/o \\ mask}}} & w/o mask & 40.67 & 90.79 & 90.91 & 90.88  \\ 
					~ & Randomly & 38.57$\scriptstyle{\pm 0.75}$ & 90.99$\scriptstyle{\pm 0.18}$ & 90.85$\scriptstyle{\pm 0.21}$ & 91.31$\scriptstyle{\pm 0.19}$  \\ \hline
					\multirow{5}{*}{\rotatebox{90}{\textbf{\makecell[c]{with \\ mask}}}} & w/o mask & 37.74 & 91.52 & 91.5 & 91.77  \\ 
					~ & Fix Low Ratio & 37.5 & 91.64 & 91.6 & 91.9  \\ 
					~ & Fix High Ratio & 36.16 & 91.44 & 91.03 & 91.9  \\ 
					~ & \textbf{Randomly} & \textbf{36.15}$\scriptstyle{\pm 0.94}$ & \textbf{91.78}$\scriptstyle{\pm 0.17}$ & \textbf{91.68}$\scriptstyle{\pm 0.24}$ &\textbf{ 92.08}$\scriptstyle{\pm 0.14}$ \\ \hline
			\end{tabular}}
	\end{table}

\noindent\textbf{\textit{Label Conditioning.}}
\label{ablation_label} As for evaluating label conditioning, we degrade the cGAN to vanilla GAN without conditions. 
Table~\ref{tab:condition} reports this ablation study. We can see that our conditioning mechanism outperforms the unconditioned scheme by a large margin. As analyzed in Sec 3.2, the conditioning spotlights the semantic discrepancy between In-D and OOD to facilitate OOD detection. 
\begin{table}[tpb]
	\centering
	\caption{Ablation study on label conditioning}
	\label{tab:condition}
	\renewcommand\arraystretch{0.9} 
	\setlength{\tabcolsep}{1mm}{\begin{tabular}{@{}cccccc@{}}
	\hline
	In-Data            & Methods     & FPR@95 ↓      & AUROC ↑      & AUPR In ↑    & AUPR Out ↑   \\ \hline
	Cifar10 In-D & \textbf{ours}        & \textbf{41.95}      & \textbf{89.57}      & \textbf{89.30 }      & \textbf{90.16}      \\
	Cifar100 OOD & uncond. & 86.94  & 63.62 & 71.26& 63.01 \\ \hline
	Cifar100 In-D & \textbf{ours}        & \textbf{50.17}      & \textbf{87.76}      & \textbf{88.18}      & \textbf{87.79}      \\
	Cifar10 OOD & uncond. & 96.74 & 52.48 & 56.29& 71.24 \\ \hline
	\end{tabular}}

	\end{table}
	

\section{Conclusion}
\label{sec:conclusion}

In this paper, we propose a novel plug-and-play OOD detection method for image classifiers, \sood, wherein we consider the semantic mismatch under masking as the distance metric.
\sood naturally learns the semantic information from the in-distribution data with the proposed mask and conditional synthesis framework. Experimental results demonstrate significantly better OOD detection capabilities of \sood over SOTA solutions.  
\\
\\
\noindent\textbf{Acknowledgements.} We appreciate the reviewers for their thoughtful comments and efforts towards improving this paper. This work was supported in part by General Research Fund of Hong Kong Research Grants Council (RGC) under Grant No. 14203521 and No. 14205420.

\clearpage
%
%
\bibliographystyle{splncs04}
\bibliography{ref.bib}

\begin{thebibliography}{10}
\providecommand{\url}[1]{\texttt{#1}}
\providecommand{\urlprefix}{URL }
\providecommand{\doi}[1]{https://doi.org/#1}

\bibitem{amodei2016concrete}
Amodei, D., Olah, C., Steinhardt, J., Christiano, P., Schulman, J., Man{\'e},
  D.: Concrete problems in ai safety. arXiv preprint arXiv:1606.06565  (2016)

\bibitem{brock2018large}
Brock, A., Donahue, J., Simonyan, K.: Large scale gan training for high
  fidelity natural image synthesis. In: International Conference on Learning
  Representations (2018)

\bibitem{choi2018waic}
Choi, H., Jang, E., Alemi, A.A.: Waic, but why? generative ensembles for robust
  anomaly detection. arXiv preprint arXiv:1810.01392  (2018)

\bibitem{cimpoi14describing}
Cimpoi, M., Maji, S., Kokkinos, I., Mohamed, S., , Vedaldi, A.: Describing
  textures in the wild. In: Proceedings of the {IEEE} Conf. on Computer Vision
  and Pattern Recognition ({CVPR}) (2014)

\bibitem{de2017modulating}
De~Vries, H., Strub, F., Mary, J., Larochelle, H., Pietquin, O., Courville,
  A.C.: Modulating early visual processing by language. Advances in Neural
  Information Processing Systems  \textbf{30} (2017)

\bibitem{taylor2018improving}
Denouden, T., Salay, R., Czarnecki, K., Abdelzad, V., Phan, B., Vernekar, S.:
  Improving reconstruction autoencoder out-of-distribution detection with
  mahalanobis distance. arXiv preprint arXiv:1812.02765  (2018)

\bibitem{dietterich2017steps}
Dietterich, T.G.: Steps toward robust artificial intelligence. AI Magazine
  \textbf{38}(3),  3--24 (2017)

\bibitem{ding2020image}
Ding, K., Ma, K., Wang, S., Simoncelli, E.P.: Image quality assessment:
  Unifying structure and texture similarity. IEEE Transactions on Pattern
  Analysis and Machine Intelligence  (2020)

\bibitem{drummond2006open}
Drummond, N., Shearer, R.: The open world assumption. In: eSI Workshop: The
  Closed World of Databases meets the Open World of the Semantic Web. vol.~15
  (2006)

\bibitem{ge2017generative}
Ge, Z., Demyanov, S., Chen, Z., Garnavi, R.: Generative openmax for multi-class
  open set classification. In: British Machine Vision Conference 2017. British
  Machine Vision Association and Society for Pattern Recognition (2017)

\bibitem{guo2021conditional}
Guo, Y., Camporese, G., Yang, W., Sperduti, A., Ballan, L.: Conditional
  variational capsule network for open set recognition. In: Proceedings of the
  IEEE/CVF International Conference on Computer Vision. pp. 103--111 (2021)

\bibitem{he2021masked}
He, K., Chen, X., Xie, S., Li, Y., Doll{\'a}r, P., Girshick, R.: Masked
  autoencoders are scalable vision learners. arXiv preprint arXiv:2111.06377
  (2021)

\bibitem{he2015delving}
He, K., Zhang, X., Ren, S., Sun, J.: Delving deep into rectifiers: Surpassing
  human-level performance on imagenet classification. In: Proceedings of the
  IEEE International Conference on Computer Vision. pp. 1026--1034 (2015)

\bibitem{he2016deep}
He, K., Zhang, X., Ren, S., Sun, J.: Deep residual learning for image
  recognition. In: Proceedings of the IEEE Conference on Computer Vision and
  Pattern Recognition. pp. 770--778 (2016)

\bibitem{hein2019relu}
Hein, M., Andriushchenko, M., Bitterwolf, J.: Why relu networks yield
  high-confidence predictions far away from the training data and how to
  mitigate the problem. In: Proceedings of the IEEE/CVF Conference on Computer
  Vision and Pattern Recognition. pp. 41--50 (2019)

\bibitem{hendrycks2016baseline}
Hendrycks, D., Gimpel, K.: A baseline for detecting misclassified and
  out-of-distribution examples in neural networks. In: 5th International
  Conference on Learning Representations (ICLR) (2017)

\bibitem{hendrycks2018deep}
Hendrycks, D., Mazeika, M., Dietterich, T.: Deep anomaly detection with outlier
  exposure. In: International Conference on Learning Representations (2018)

\bibitem{8721151}
Huang, G., Liu, Z., Pleiss, G., Van Der~Maaten, L., Weinberger, K.:
  Convolutional networks with dense connectivity. IEEE Transactions on Pattern
  Analysis and Machine Intelligence pp.~1--1 (2019)

\bibitem{huang2020feature}
Huang, H., Li, Z., Wang, L., Chen, S., Zhou, X., Dong, B.: Feature space
  singularity for out-of-distribution detection. In: Proceedings of the
  Workshop on Artificial Intelligence Safety (SafeAI) (2021)

\bibitem{huang2021importance}
Huang, R., Geng, A., Li, Y.: On the importance of gradients for detecting
  distributional shifts in the wild. Advances in Neural Information Processing
  Systems  \textbf{34} (2021)

\bibitem{kingma2013auto}
Kingma, D.P., Welling, M.: Auto-encoding variational bayes. In: 2nd
  International Conference on Learning Representations (ICLR) (2014)

\bibitem{kirichenko2020normalizing}
Kirichenko, P., Izmailov, P., Wilson, A.G.: Why normalizing flows fail to
  detect out-of-distribution data. Advances in neural information processing
  systems  \textbf{33},  20578--20589 (2020)

\bibitem{krizhevsky2009learning}
Krizhevsky, A., Hinton, G., et~al.: Learning multiple layers of features from
  tiny images  (2009)

\bibitem{krizhevskyimagenet}
Krizhevsky, A., Sutskever, I., Hinton, G.E.: Imagenet classification with deep
  convolutional neural networks. In: Advances in Neural Information Processing
  Systems. vol.~25, pp. 1106--1114 (2012)

\bibitem{tiny}
Le, Y., Yang, X.: Tiny imagenet visual recognition challenge. CS 231N
  \textbf{7}(7), ~3 (2015)

\bibitem{lee2018simple}
Lee, K., Lee, K., Lee, H., Shin, J.: A simple unified framework for detecting
  out-of-distribution samples and adversarial attacks. Advances in neural
  information processing systems  \textbf{31} (2018)

\bibitem{liang2018enhancing}
Liang, S., Li, Y., Srikant, R.: Enhancing the reliability of
  out-of-distribution image detection in neural networks. In: 6th International
  Conference on Learning Representations, ICLR 2018 (2018)

\bibitem{lin2021mood}
Lin, Z., Roy, S.D., Li, Y.: Mood: Multi-level out-of-distribution detection.
  In: Proceedings of the IEEE/CVF Conference on Computer Vision and Pattern
  Recognition. pp. 15313--15323 (2021)

\bibitem{EBO}
Liu, W., Wang, X., Owens, J., Li, Y.: Energy-based out-of-distribution
  detection. In: Advances in Neural Information Processing Systems. vol.~33,
  pp. 21464--21475. Curran Associates, Inc. (2020)

\bibitem{miyato2018cgans}
Miyato, T., Koyama, M.: cgans with projection discriminator. In: International
  Conference on Learning Representations (2018)

\bibitem{nalisnick2018deep}
Nalisnick, E., Matsukawa, A., Teh, Y.W., Gorur, D., Lakshminarayanan, B.: Do
  deep generative models know what they don't know? In: International
  Conference on Learning Representations (2019)

\bibitem{neal2018open}
Neal, L., Olson, M., Fern, X., Wong, W.K., Li, F.: Open set learning with
  counterfactual images. In: Proceedings of the European Conference on Computer
  Vision (ECCV). pp. 613--628 (2018)

\bibitem{37648}
Netzer, Y., Wang, T., Coates, A., Bissacco, A., Wu, B., Ng, A.Y.: Reading
  digits in natural images with unsupervised feature learning. In: NIPS
  Workshop on Deep Learning and Unsupervised Feature Learning 2011 (2011)

\bibitem{nguyen2015deep}
Nguyen, A., Yosinski, J., Clune, J.: Deep neural networks are easily fooled:
  High confidence predictions for unrecognizable images. In: Proceedings of the
  IEEE Conference on Computer Vision and Pattern Recognition. pp. 427--436
  (2015)

\bibitem{oza2019c2ae}
Oza, P., Patel, V.M.: C2ae: Class conditioned auto-encoder for open-set
  recognition. In: Proceedings of the IEEE/CVF Conference on Computer Vision
  and Pattern Recognition. pp. 2307--2316 (2019)

\bibitem{pidhorskyi2018generative}
Pidhorskyi, S., Almohsen, R., Doretto, G.: Generative probabilistic novelty
  detection with adversarial autoencoders. Advances in neural information
  processing systems  \textbf{31} (2018)

\bibitem{ren2019likelihood}
Ren, J., Liu, P.J., Fertig, E., Snoek, J., Poplin, R., Depristo, M., Dillon,
  J., Lakshminarayanan, B.: Likelihood ratios for out-of-distribution
  detection. Advances in Neural Information Processing Systems  \textbf{32}
  (2019)

\bibitem{sara2019image}
Sara, U., Akter, M., Uddin, M.S.: Image quality assessment through fsim, ssim,
  mse and psnr—a comparative study. Journal of Computer and Communications
  \textbf{7}(3),  8--18 (2019)

\bibitem{schlegl2017unsupervised}
Schlegl, T., Seeb{\"o}ck, P., Waldstein, S.M., Schmidt-Erfurth, U., Langs, G.:
  Unsupervised anomaly detection with generative adversarial networks to guide
  marker discovery. In: International conference on information processing in
  medical imaging. pp. 146--157. Springer (2017)

\bibitem{schonfeld2020u}
Schonfeld, E., Schiele, B., Khoreva, A.: A u-net based discriminator for
  generative adversarial networks. In: Proceedings of the IEEE/CVF Conference
  on Computer Vision and Pattern Recognition. pp. 8207--8216 (2020)

\bibitem{sricharan2018building}
Sricharan, K., Srivastava, A.: Building robust classifiers through generation
  of confident out of distribution examples. arXiv preprint arXiv:1812.00239
  (2018)

\bibitem{vernekar2019out}
Vernekar, S., Gaurav, A., Abdelzad, V., Denouden, T., Salay, R., Czarnecki, K.:
  Out-of-distribution detection in classifiers via generation. arXiv preprint
  arXiv:1910.04241  (2019)

\bibitem{wang2021can}
Wang, H., Liu, W., Bocchieri, A., Li, Y.: Can multi-label classification
  networks know what they don't know? Advances in Neural Information Processing
  Systems  \textbf{34} (2021)

\bibitem{wang2021energy}
Wang, Y., Li, B., Che, T., Zhou, K., Liu, Z., Li, D.: Energy-based open-world
  uncertainty modeling for confidence calibration. In: Proceedings of the
  IEEE/CVF International Conference on Computer Vision. pp. 9302--9311 (2021)

\bibitem{yang2021semantically}
Yang, J., Wang, H., Feng, L., Yan, X., Zheng, H., Zhang, W., Liu, Z.:
  Semantically coherent out-of-distribution detection. In: Proceedings of the
  IEEE/CVF International Conference on Computer Vision. pp. 8301--8309 (2021)

\bibitem{yang2021generalized}
Yang, J., Zhou, K., Li, Y., Liu, Z.: Generalized out-of-distribution detection:
  A survey. arXiv preprint arXiv:2110.11334  (2021)

\bibitem{contraNet}
Yang, Y., Gao, R., Li, Y., Lai, Q., Xu, Q.: What you see is not what the
  network infers: Detecting adversarial examples based on semantic
  contradiction. In: Network and Distributed System Security Symposium (NDSS)
  (2022)

\bibitem{yu2016lsun}
Yu, F., Seff, A., Zhang, Y., Song, S., Funkhouser, T., Xiao, J.: Lsun:
  Construction of a large-scale image dataset using deep learning with humans
  in the loop (2016)

\bibitem{Yu_2019_ICCV}
Yu, Q., Aizawa, K.: Unsupervised out-of-distribution detection by maximum
  classifier discrepancy. In: Proceedings of the IEEE/CVF International
  Conference on Computer Vision (ICCV) (2019)

\bibitem{zaeemzadeh2021out}
Zaeemzadeh, A., Bisagno, N., Sambugaro, Z., Conci, N., Rahnavard, N., Shah, M.:
  Out-of-distribution detection using union of 1-dimensional subspaces. In:
  Proceedings of the IEEE/CVF Conference on Computer Vision and Pattern
  Recognition. pp. 9452--9461 (2021)

\bibitem{zagoruyko2016wide}
Zagoruyko, S., Komodakis, N.: Wide residual networks. In: British Machine
  Vision Conference 2016. British Machine Vision Association (2016)

\bibitem{zhang2019self}
Zhang, H., Goodfellow, I., Metaxas, D., Odena, A.: Self-attention generative
  adversarial networks. In: International Conference on Machine Learning (2019)

\bibitem{zhang2018unreasonable}
Zhang, R., Isola, P., Efros, A.A., Shechtman, E., Wang, O.: The unreasonable
  effectiveness of deep features as a perceptual metric. In: Proceedings of the
  IEEE Conference on Computer Vision and Pattern Recognition. pp. 586--595
  (2018)

\bibitem{zhou2017places}
Zhou, B., Lapedriza, A., Khosla, A., Oliva, A., Torralba, A.: Places: A 10
  million image database for scene recognition. IEEE Transactions on Pattern
  Analysis and Machine Intelligence  (2017)

\end{thebibliography}
\appendix
\noindent {\Large\textbf{Appendix}}

\renewcommand{\thesection}{\Alph{section}}
\section{Training Process of Generative Model}
\subsection{Objective Functions}
We implement the adversarial loss with a U-net based discriminator~\cite{schonfeld2020u}, denoted as $D^{Unet}$.
$D^{Unet}$ contains two components: $D^{Unet}_{enc}$ and $D^{Unet}_{dec}$.
$D^{Unet}_{enc}\in\mathbb{R}$ provides the real/fake decision as a scalar.
While $D^{Unet}_{dec}\in\mathbb{R}^{I}$ generates a per-pixel real/fake map for the input image, where $I=h\times w$ indicates the scale of input image.
Compared to the vanilla discriminator, $D^{Unet}$ not only determines whether the input image is realistic or fake, but also tries to locate the fake parts.
Empowered by the per-pixel real/fake map, our generative model can be optimized to focus more on structural semantic features and synthesize coherent image both globally and locally as desired.
We formulate the adversarial loss for the discriminator in \equ{\ref{eq:dis_u}}-\equ{\ref{eq:dis_dec}}:
\begin{gather}
	\mathcal{L}_{D^{Unet}} = \mathcal{L}_{D^{Unet}_{enc}} + \mathcal{L}_{D^{Unet}_{dec}}\text{,}\label{eq:dis_u}\\
\mathcal{L}_{D^{Unet}_{enc}}= -\mathbb{E}_x[\log D^{Unet}_{enc}(x,y)]- \mathbb{E}_x[\log (1-D^{Unet}_{enc}(x',y))]\text{,}	\label{eq:dis_en}\\
\mathcal{L}_{D^{Unet}_{dec}} =  -\mathbb{E}_x\Big[\sum_{I}\log D^{Unet}_{dec}(x,y)\Big] - \mathbb{E}_x\Big[\sum_{I}\log (1-D^{Unet}_{dec}(x',y))\Big]\text{,}\label{eq:dis_dec}
\end{gather}
where $\mathcal{L}_{D^{Unet}}$ and $\mathcal{L}_{D^{Unet}_{dec}}$ are the loss functions for $D^{Unet}_{enc}$ and $D^{Unet}_{dec}$, respectively.
Correspondingly, the adversarial loss applied on the generator is as follow:
\begin{small}
\begin{equation}
	\label{eq:g}
	\mathcal{L}_{\mathbf{G}} = -\mathbb{E}_x\Big[\log D^{Unet}_{enc}(x',y)  \\
	+ \sum_{I}\log D^{Unet}_{dec}(x')\Big] + \ell_{1}(x,x') +\ell_{2}(x,x') + \mathcal{SSIM}(x,x'). 
\end{equation}
\end{small}

\subsection{Training Process}
\subsubsection{\textit{Encoder.}}
We adopt a four-layer convolutional neural network as the feature extractor for Encoder, then two fully-connected layers are employed to output $\mu$ and $\varSigma$.
The dimension of the latent variable $z$ is set at 128.

\subsubsection{\textit{Decoder.}}
We employ the generator architecture proposed in \cite{brock2018large} as our Decoder's backbone, then reset the input size to (3, 32, 32), and the channel multiplier to 32, which represents the number of units in each layer\cite{brock2018large}.
The input latent variable size equals 128.

\subsubsection{\textit{Discriminator.}}
We build $D^{Unet}$ based on the implementation of \cite{schonfeld2020u}, changing the channel multiplier to 32.

All three models mentioned above are trained from scratches in an end-to-end way.
We use Adam~\cite{kingma2015adam} as the optimizer, with $\beta_1 = 0$, $\beta_2=0.999$, learning rate fixed at $5\cdot10^{-5}$. The batch size is set at 96.
We detail the training process of our generative model in Algorithm~\ref{alg:sood}.

\begin{algorithm}[hbt]
	\caption{Training Framework of \G}\label{alg:sood}
	\SetKwInOut{Input}{Input}
	\SetKwInOut{Output}{Output}

	\Input{Training data $\mathcal{X} = \{x\}^N$, $\mathcal{Y}=\{y\}^N$, the random mask $\mathbf{M}$}
	\Output{The parameters of $\mathbf{E}$, $\mathbf{D}$}
	
	\For {some training iterations}{
		$x' = \mathbf{G}(\mathbf{M}(x),y) = \mathbf{D}(\mathbf{E}(\mathbf{M}(x), y))$\;\
		Feed ($x$, $y$) and ($x'$, $y$) into $D^{Unet}$, {\color{red}respectively}\;\
		Optimize {$\mathbf{D}$} and $\mathbf{E}$ for $\mathcal{L}_{\mathbf{G}}$(Eq.~\eqref{eq:g}) and $\mathcal{L}_{KLD}$\;\ 
		Optimize {$D^{Unet}$} for $\mathcal{L}_{D^{Unet}}$ (Eq.~\eqref{eq:dis_u})\;
	}
	\Return{$\mathbf{E}$, $\mathbf{G}$}
\end{algorithm}

\section{Quantitative Results}
In this section, we provide more experimental results on \cifar and \cifarh benchmarks, respectively. Furthermore, to further validate the effectiveness of the proposed \textit{conditional binary classifier} ($\mathcal{C}_b$) in the anomalous scoring model, we detail its performance in each OOD dataset by varying the type of $\mathcal{C}_b$, i.e. trained with/without external OOD data.

\subsection{More Results on \cifar Benchmarks}
Table~\ref{tab:cifar10} presents the comparison of our \sood trained with external unlabeled data sourced from \tinyimagenet, and baselines implemented with extra data.  We conclude that \sood  outperforms or at least on par with baselines on \cifar benchmarks. 

Additionally, in Table~\ref{tab:cifar10} we observe that OE and UDG achieve a much better  performance on \svhn than on other OOD datasets.
In fact, most street number images contained in \svhn have relatively flat backgrounds, as shown in Fig.~\ref{fig:OODcifar10} and Fig.~\ref{fig:OODcifar100}'s \svhn columns. In this case, OE and UDG can achieve excellent performance by overfitting to this specific low-level feature of \svhn instead of considering the semantic level change caused by \svhn. Thus, when encountering a more challenging case, e.g., \cifarh, which has the same data source as \cifar but different semantic meanings, both OE and UDG suffer a noticeable performance degradation. In contrast, \sood identifies OOD according to their semantic mismatch, thus, remains stable performance on various OODs.  

\begin{table}[H]
    \caption{OOD Detection Performance on \cifar benchmarks, \sood trained with external OOD data. All the values are in percentages. $\uparrow$/$\downarrow$ indicates higher/lower value is better. The best results are in \textbf{bold}.}
	\label{tab:cifar10}
	\fontsize{6}{8}\selectfont
	\begin{tabularx}{\textwidth}{ P{1.5cm} P{2cm} s s s s P{1.7cm} }
	\toprule 
	
	\begin{tabular}{@{}c@{}}\bf{Detection} \\ \bf{Methods} \end{tabular}
	& \begin{tabular}{@{}c@{}}\bf{OOD} \end{tabular}
	& \begin{tabular}{@{}c@{}} \bf{FPR@} \\ \bf{TPR95\%} \\ \bf{$\downarrow$} \end{tabular} 
	& \begin{tabular}{@{}c@{}} \bf{AUROC}   \\ \bf{$\uparrow$} \end{tabular}
	& \begin{tabular}{@{}c@{}} \bf{AUPR} \\ \bf{In} \\ $\uparrow$ \end{tabular}
	&  \begin{tabular}{@{}c@{}} \bf{AUPR} \\ \bf{Out} \\ $\uparrow$ \end{tabular}
	&  \begin{tabular}{@{}c@{}} \bf{Classification} \\ \bf{Accuracy} \\ $\uparrow$ \end{tabular}\\ 
    \hline
	\multirow{7}{*}{\bf{MCD}}
	& \svhn &60.27 &89.78 &85.33 &94.25 &90.56 \\
	& \cifarh &74.00 &82.78 &83.97 &79.16 &90.56 \\
	& \tinyimagenet &78.89 &80.98 &85.63 &72.48 &87.33 \\
	& \texture &83.92 &81.59 &90.20 &63.27 &90.56  \\
	& \lsun &68.96 &84.71 &85.74 &81.50 &90.56 \\
	& \places &72.08 &83.51 &69.44 &92.52 &88.51 \\\cline{2-7}
	& \textbf{Mean}&73.02 &83.89 &83.39 &80.53 &89.68 \\
	\hline

	\multirow{7}{*}{\bf{OE}}
	& \svhn &20.88 &96.43 &93.62& 98.32&91.87\\
	& \cifarh &58.54&86.22 &86.17 &84.88 &91.87  \\
	& \tinyimagenet &58.98 &87.65 &90.09 & 82.16&89.27  \\
	& \texture &51.17 &89.56 &93.79 &81.88 &91.87  \\
	& \lsun &57.97 &86.75 &87.69 &85.07 &91.87\\
	& \places &55.64 &87.00 &73.11 &94.67 &90.99  \\\cline{2-7}
	& \textbf{Mean}&50.53 &88.93 &87.55 &87.83 &91.29 \\
	 \hline

	\multirow{7}{*}{\bf{UDG}}
	& \svhn &\textbf{13.26} &\textbf{97.49} &\textbf{95.66} &\textbf{98.69} &92.94 \\
	& \cifarh &47.20&90.98 &\textbf{91.74} &89.36 & 92.94\\
	& \tinyimagenet &50.18 &91.91 &\textbf{94.43} &86.99 &90.22 \\
	& \texture &20.43 &96.44 &98.12 &92.91 &92.94 \\
	& \lsun &42.05 &93.21 &94.53 &91.03 &92.94 \\
	& \places &44.22 &92.64 &87.17 &96.66 &91.68 \\\cline{2-7}
	& \textbf{Mean} &36.22 &93.78 &93.61 &92.61 &92.28 \\
	 
	\hline

	\multirow{7}{*}{\textbf{Ours}}
	& \svhn & 24.27 &95.93 &92.98 &98.05 &\textbf{95.02} \\
	& \cifarh &\textbf{39.92} &\textbf{91.45} & 91.54 &\textbf{91.73} &\textbf{95.02} \\
	& \tinyimagenet &\textbf{32.41} &\textbf{93.34} &93.63 &\textbf{93.41} &\textbf{92.54} \\
	& \texture &\textbf{6.86} &\textbf{98.69} &\textbf{99.29} &\textbf{97.71} &\textbf{95.02} \\
	& \lsun &\textbf{33.31} &\textbf{93.40} &\textbf{93.85} &\textbf{93.22} & \textbf{95.02}\\
	& \places & \textbf{35.51} &\textbf{92.77} &\textbf{82.25} &\textbf{94.82} & \textbf{93.87}\\\cline{2-7}
	& \textbf{Mean}&\textbf{28.71} &\textbf{94.27} & \textbf{92.26} &\textbf{94.82} & \textbf{94.42}\\	
	\bottomrule
	\end{tabularx}
\end{table}

\begin{table}[tpb]
    \centering
    \caption{OOD Detection Performance on \cifarh as In-D, \sood training without external data. All the values are in percentages. $\uparrow$/$\downarrow$ indicates higher/lower value is better. The best results are in \textbf{bold}.} 
    \label{tab:cifar100}

    \fontsize{6}{8}\selectfont
	\begin{tabularx}{\textwidth}{ P{1.5cm} P{2cm} s s s s P{1.7cm} }
	\toprule 
	
	\begin{tabular}{@{}c@{}} \bf{Detection} \\ \bf{Methods} \end{tabular}
	& \begin{tabular}{@{}c@{}} \bf{OOD} \end{tabular}
	& \begin{tabular}{@{}c@{}} \bf{FPR@} \\ \bf{TPR95\%} \\ \bf{$\downarrow$} \end{tabular} 
	& \begin{tabular}{@{}c@{}} \bf{AUROC}  \\ \bf{$\uparrow$} \end{tabular}
	& \begin{tabular}{@{}c@{}} \bf{AUPR} \\ \bf{In} \\ $\uparrow$ \end{tabular}
	&  \begin{tabular}{@{}c@{}} \bf{AUPR} \\ \bf{Out} \\ $\uparrow$ \end{tabular}
	&  \begin{tabular}{@{}c@{}} \bf{Classification} \\ \bf{Accuracy} \\ $\uparrow$ \end{tabular}\\

	 \hline

	\multirow{7}{*}{\bf{ODIN}}
	 & \svhn &90.33 &75.59 &65.25 &84.49 &76.65 \\
	 & \cifar &81.28 &77.90 &79.93 &73.39 &76.65 \\
	 & \tinyimagenet &82.74 &77.58 &86.26 &61.38 &69.56 \\
	 & \texture &79.47 &77.92 &86.69 &62.97 &76.65 \\
	 & \lsun &80.57 &78.22 &86.34 &63.44 &76.10 \\
	 & \places &76.42 &80.66 &66.77 &89.66 &77.56 \\\cline{2-7}
	 & \textbf{Mean}&81.89 &77.98 &78.54 &72.56 &75.53 \\
	 
	\hline
	\multirow{7}{*}{\bf{EBO}}
	& \svhn &78.23 &83.57 &75.61 &90.24 &76.65 \\
	& \cifar &81.25 &78.95 &80.01 &74.44 &76.65 \\
	& \tinyimagenet &83.32 &78.34 &87.08 &62.13 &69.56 \\
	& \texture &84.29 &76.32 &85.87 &59.12 &76.65 \\
	& \lsun &84.51 &77.66 &86.42 &61.40 &76.10 \\
	& \places &78.37 &80.99 &68.22 &89.60 &77.56 \\\cline{2-7}
	& \textbf{Mean}& 81.66 & 79.31& 80.54& 72.82 &75.53 \\
	 
	\hline
	\multirow{7}{*}{\bf{Ours}}
	& \svhn & \textbf{58.16} &\textbf{87.38} &\textbf{78.25}& \textbf{93.81}& \textbf{76.65}\\
	& \cifar &\textbf{54.31} &\textbf{85.91}&\textbf{86.27} &\textbf{85.91}& \textbf{76.65}\\
	& \tinyimagenet & \textbf{55.33}&\textbf{86.95} &\textbf{87.55} &\textbf{86.67} & \textbf{69.56}\\
	& \texture & \textbf{46.70}&\textbf{89.20} &\textbf{93.48} &\textbf{83.28} & \textbf{76.65}\\
	& \lsun & \textbf{53.43}&\textbf{87.98}& \textbf{88.82}&\textbf{87.32}& \textbf{76.10}\\
	& \places &\textbf{54.20}&\textbf{87.41} &\textbf{71.68} &\textbf{95.78}& \textbf{77.56}\\\cline{2-7}
	& \textbf{Mean}& \textbf{53.69} &\textbf{87.47}& \textbf{84.34} & \textbf{88.80} &\textbf{75.53} \\
	\bottomrule
	\end{tabularx}
\end{table}

\subsection{More Results on \cifarh benchmarks}

Table~\ref{tab:cifar100} shows the comparison of our \sood trained without external OOD data, and baselines are implemented under the same setting. We conclude that \sood achieves state-of-the-art performance on \cifarh benchmarks. 

\subsection{Ablation Study on Conditional Binary Classifier}

To study how much the proposed \textit{Conditional Binary Classifier} ($\mathcal{C}_b$) contributes to \sood, we conduct several ablations on $\mathcal{C}_b$. More specifically, we consider three configurations: $\mathcal{C}_b$, $\mathcal{C}_b$(\tinyimagenet), and $\mathcal{C}_b+$ $\mathcal{C}_b$(\tinyimagenet), where $\mathcal{C}_b$ referring to the Conditional Binary Classifier trained only on In-D samples, $\mathcal{C}_b$(\tinyimagenet) denoted the Conditional Binary Classifier using \tinyimagenet as extra training data and $\mathcal{C}_b+$ $\mathcal{C}_b$(\tinyimagenet) indicating that $\mathcal{C}_b$ and $\mathcal{C}_b$ (\tinyimagenet) are used in a cascade way. 

Table~\ref{tab:ablation_cifar10} and Table~\ref{tab:ablation_cifar100} demonstrate $\mathcal{C}_b$'s performance on \cifar and \cifarh benchmarks in six OOD datasets, respectively. The main takeaways are: \textbf{(1)} $\mathcal{C}_b$ or $\mathcal{C}_b$(\tinyimagenet)  alone can achieve acceptable performance; \textbf{(2)}  $\mathcal{C}_b$(\tinyimagenet) outperforms $\mathcal{C}_b$, which means that adding external unlabeled data into the training process can improve the detection ability; \textbf{(3)} coupling scorers, here $\mathcal{C}_b+$ $\mathcal{C}_b$(\tinyimagenet), usually leads to a better detection capability than that of any single scorer within the coupling. Above findings align with what we have reported in our paper,  and further indicate that $\mathcal{C}_b$ plays a key role in the proposed anomalous scoring model.

\begin{table}[htb]

    \centering
    \caption{Conditional Binary Classifier Performance on \cifar benchmarks. All the values are in percentages. $\uparrow$/$\downarrow$ indicates higher/lower value is better. The best results are in \textbf{bold}.
    $\mathcal{C}_b$ and $\mathcal{C}_b$(\tinyimagenet) indicates the proposed model trained without/with external unlabeled \tinyimagenet data, respectively. }
    \label{tab:ablation_cifar10}

    \fontsize{6}{8}\selectfont
	\begin{tabularx}{\textwidth}{ P{2.5cm} P{2cm} s s s s }
	\toprule 
	
	\begin{tabular}{@{}c@{}} \bf{Anomalous} \\ \bf{Scoring Model} \end{tabular}
	& \begin{tabular}{@{}c@{}} \bf{OOD} \end{tabular}
	& \begin{tabular}{@{}c@{}} \bf{FPR@} \\ \bf{TPR95\%} \\ \bf{$\downarrow$} \end{tabular} 
	& \begin{tabular}{@{}c@{}} \bf{AUROC}  \\ \bf{$\uparrow$} \end{tabular}
	& \begin{tabular}{@{}c@{}} \bf{AUPR} \\ \bf{In} \\ $\uparrow$ \end{tabular}
	&  \begin{tabular}{@{}c@{}} \bf{AUPR} \\ \bf{Out} \\ $\uparrow$ \end{tabular}\\

	 \hline
	\multirow{7}{*}{$\mathcal{C}_b$}
  & \svhn                  & 48.01 & 86.85 & 75.20 & 94.34 \\
  & \cifarh                & 42.80 & 89.13 & 88.58 & 89.85 \\
  & \tinyimagenet          & 40.54 & 89.78 & 89.27 & 90.42 \\
  & \texture               & 42.54 & 87.47 & 91.33 & 83.85 \\
  & \lsun                  & 43.76 & 90.15 & 90.18 & 90.17 \\
  & \places                & 43.49 & 89.40 & 72.82 & 96.65 \\ \cline{2-6}
  & \textbf{Mean}          & 43.52  & 88.80  & 84.56  & 90.88 \\ 
	 
	\hline
	\multirow{7}{*}{\makecell[c]{$\mathcal{C}_b$ (\tinyimagenet)}}
  & \svhn                 & 39.47 & 91.49 & 83.58 & 96.23 \\
  & \cifarh               & 37.43 & 91.31 & 91.02 & 91.73 \\
  & \tinyimagenet         & 31.92 & 93.10 & 93.01 & 93.34 \\
  & \texture              & 25.74 & 94.25 & 96.17 & 91.89 \\
  & \lsun                 & 32.74 & 93.55 & 93.83 & 93.40 \\
  & \places               & 34.45 & 92.78 & 81.48 & 97.71 \\\cline{2-6}
  & \textbf{Mean}         & 33.63  & 92.75  & 89.85  & 94.05 \\
	 
	\hline
	\multirow{7}{*}{\makecell[c]{$\mathcal{C}_b+$ \\ $\mathcal{C}_b$ (\tinyimagenet)}}
  & \svhn                  & \textbf{39.44} & \textbf{91.50} & \textbf{83.60} &\textbf{96.25} \\
  & \cifarh                & \textbf{36.64} & \textbf{91.40} & \textbf{91.15} & \textbf{91.85} \\
  & \tinyimagenet          & \textbf{31.86} & \textbf{93.12} & \textbf{93.04} & \textbf{93.38} \\
  & \texture               & \textbf{25.37} & \textbf{94.34} & \textbf{96.24} & \textbf{92.00} \\
  & \lsun                  & \textbf{32.67} & \textbf{93.55}& \textbf{93.84} &\textbf{ 93.41} \\
  & \places                &\textbf{34.42} & \textbf{92.79} & \textbf{81.51} & \textbf{97.72} \\ \cline{2-6} 
  & \textbf{Mean}          & \textbf{33.40}  & \textbf{92.78}  & \textbf{89.90}  &\textbf{94.10} \\
	\bottomrule
	\end{tabularx}
\end{table}

\begin{table}[htb]
    \centering
    \caption{Conditional Binary Classifier Performance on \cifarh benchmarks. All the values are in percentages. $\uparrow$/$\downarrow$ indicates higher/lower value is better. The best results are in \textbf{bold}. $\mathcal{C}_b$ and $\mathcal{C}_b$(\tinyimagenet) indicates the proposed model trained without/with external unlabeled \tinyimagenet data, respectively.} 
    \label{tab:ablation_cifar100}
    
    \fontsize{6}{9}\selectfont
	\begin{tabularx}{\textwidth}{ P{2.5cm} P{2cm} s s s s }
	\toprule 
	
	\begin{tabular}{@{}c@{}} \bf{Anomalous} \\ \bf{Scoring Model} \end{tabular}
	& \begin{tabular}{@{}c@{}} \bf{OOD} \end{tabular}
	& \begin{tabular}{@{}c@{}} \bf{FPR@} \\ \bf{TPR95\%} \\ \bf{$\downarrow$} \end{tabular} 
	& \begin{tabular}{@{}c@{}} \bf{AUROC}  \\ \bf{$\uparrow$} \end{tabular}
	& \begin{tabular}{@{}c@{}} \bf{AUPR} \\ \bf{In} \\ $\uparrow$ \end{tabular}
	&  \begin{tabular}{@{}c@{}} \bf{AUPR} \\ \bf{Out} \\ $\uparrow$ \end{tabular}\\

	 \hline
	\multirow{7}{*}{$\mathcal{C}_b$}
  & \svhn                   & 65.18 & 81.32 & 65.61 & 91.35 \\
  & \cifar                  & 55.11 & 85.75 & 85.78 & 85.99 \\
  & \tinyimagenet           & 54.69 & 86.27 & 86.26 & 86.43 \\
  & \texture                & 56.63 & 83.30 & 88.40 & 77.17 \\
  & \lsun                   & 54.77 & 86.96 & 87.20 & 86.83 \\
  & \places                 & 54.18 & 86.36 & 67.60 & 95.54 \\\cline{2-6}
  & \textbf{Mean}           & 56.76 & 84.99 & 80.14 & 87.22 \\
	 
	\hline
	\multirow{7}{*}{\makecell[c]{$\mathcal{C}_b$ (\tinyimagenet)}}
  & \svhn                     & 54.61 & \textbf{86.30} & \textbf{74.30} & 93.80 \\
  & \cifar                    & 49.82 & 87.57 & 87.74 & 87.69 \\
  & \tinyimagenet             & 45.86 & 89.38 & 89.48 & 89.43 \\
  & \texture                  & 48.24 & 87.16 & 91.55 & 81.83 \\
  & \lsun                     & 44.43 & 90.07 & 90.25 & 90.00 \\
  & \places                   & 46.89 & 88.93 & 72.99 & 96.41 \\\cline{2-6}
  & \textbf{Mean}             & 48.31  & 88.24  & 84.39  & 89.86 \\
	 
	\hline
	\multirow{7}{*}{\makecell[c]{$\mathcal{C}_b+$ \\ $\mathcal{C}_b$ (\tinyimagenet)}}
  & \svhn                 & \textbf{54.31} & \textbf{86.30} & \textbf{74.30} & \textbf{93.81} \\         
  & \cifar                & \textbf{49.62} & \textbf{87.60} & \textbf{87.77} & \textbf{87.77} \\         
  & \tinyimagenet         & \textbf{45.46} & \textbf{89.39} & \textbf{89.48} & \textbf{89.48} \\  
  & \texture              & \textbf{47.18} & \textbf{87.37} & \textbf{91.71} & \textbf{82.17} \\         
  & \lsun                 & \textbf{44.01} & \textbf{90.08} & \textbf{90.26} & \textbf{90.04} \\         
  & \places               & \textbf{46.73} & \textbf{88.95} & \textbf{73.02} &\textbf{ 96.43} \\ \cline{2-6}        
  & \textbf{Mean}         & \textbf{47.89}  & \textbf{88.28}  & \textbf{84.42}  & \textbf{89.95}  \\   
	\bottomrule
	\end{tabularx}
\end{table}

\subsection{Ablation Study on Masking Style}
\begin{figure}[tbp]
	\centering\includegraphics[width=\linewidth]{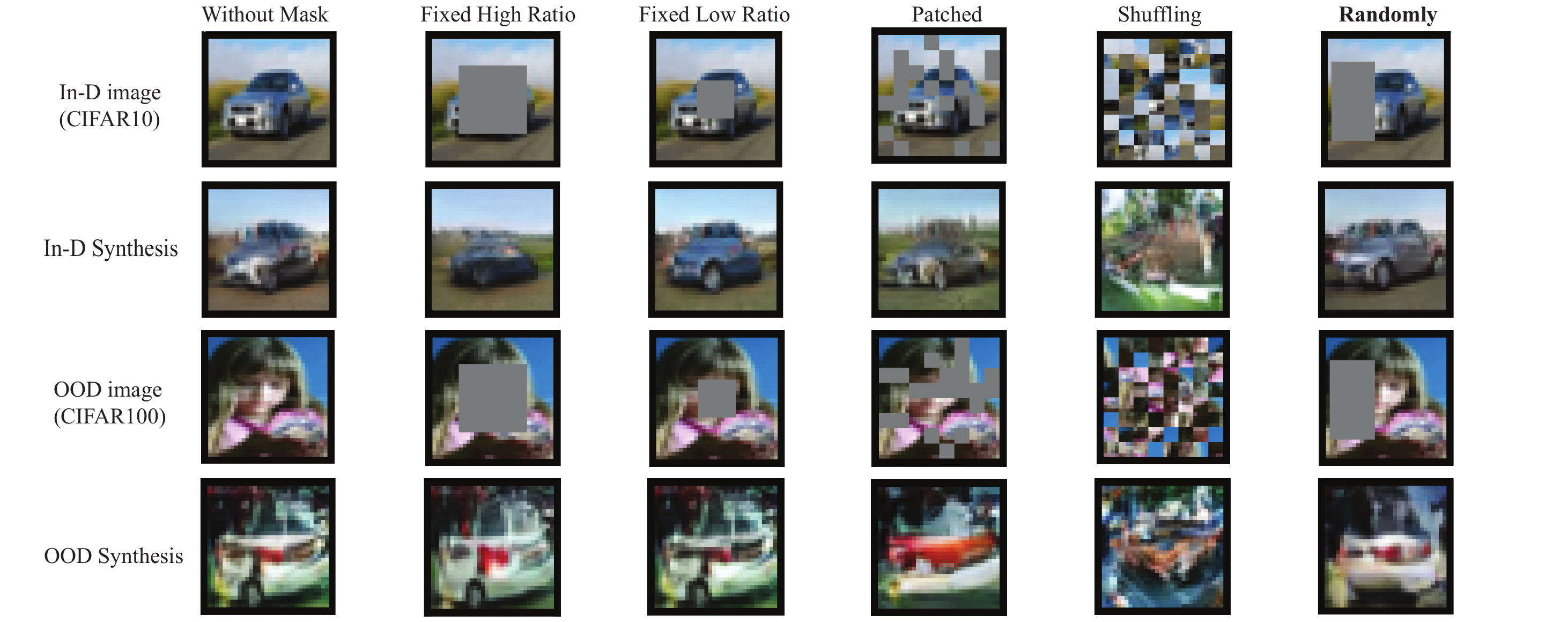}

	\caption{Visualization of different masking styles and their impacts on synthesized images. The semantic label is assigned as ``car'' for both the In-D image and the OOD image. We set the masking ratio as 0.3 for ``Fixed High Ratio'' and ``Patched'', 0.1 for ``Fixed Low Ratio'', and that of ``Randomly'' varies from 0.1 to 0.3. \sood employs the \textbf{Randomly} masking style. }
	\label{fig:masking}

\end{figure}

We try several masking forms as exemplified in Fig.~\ref{fig:masking}, and summarize the corresponding experimental results in Table~\ref{tab:mask}.
Experiments show the randomly masking outperforms other strategies.

From the first three rows in Table~\ref{tab:mask}, we notice that masking can indeed help with performance improvement.
However, as we can observe from the second column in Fig.~\ref{fig:masking}, a fixed mask with high ratio (e.g., 0.3) can lead the synthesis to loss of fine details.
In addition, we implement a patched masking like \cite{he2021masked}.
However, such masking style may break the continuity within the image, thus lead to low quality on the synthesis for In-D.
We also try a non-masking strategy, shuffling, but it further breaks the continuity of the image.
Finally, we identify that the most effective strategy is randomly masking.
As can be seen, both the quality of the synthesis in Fig.~\ref{fig:masking} and the overall performance in Table~\ref{tab:mask} outperform other strategies.

\begin{table}[tbp]

	\centering
    \caption{Ablation studies on different masking styles. The results are obtained by setting \cifar as In-D, \cifarh as OOD, with \sood trained on extra \tinyimagenet acting as OOD. The \textbf{bolded} values are the highest performance. All the values are in percentages. $\uparrow$/$\downarrow$ indicates higher/lower value is better. }
	\label{tab:mask}
	\fontsize{6}{9}\selectfont
	\begin{tabularx}{\textwidth}{ s s s s s }
	\hline 
	
	\begin{tabular}{@{}c@{}}\bf{Mask Style} \end{tabular}
	& \begin{tabular}{@{}c@{}} \bf{FPR@TPR95\%}  \bf{$\downarrow$} \end{tabular} 
	& \begin{tabular}{@{}c@{}} \bf{AUROC}   \bf{$\uparrow$} \end{tabular}
	& \begin{tabular}{@{}c@{}} \bf{AUPR-In}  $\uparrow$ \end{tabular}
	&  \begin{tabular}{@{}c@{}} \bf{AUPR-Out}  $\uparrow$ \end{tabular}\\
    \hline 

		Without Masking     & 40.53                & 91.26          & 91.25            & 91.55             \\
		Fixed Low Ratio     & 40.20                & 91.33          & 91.32            & 91.59             \\
		Fixed High Ratio    & {\ul \textit{39.57}} & {\ul \textit{91.56}} & {\ul \textit{91.48}} & {\ul \textit{91.87}} \\
		Patched             & 39.81                & 91.34          & 91.33            & 91.64             \\
		Shuffling           & 44.14                & 88.73          & 88.15            & 89.18             \\
		\textbf{Randomly}   & \textbf{39.48}       & \textbf{91.66} & \textbf{91.65}   & \textbf{91.95}    \\ 
		\hline
	\end{tabularx}

\end{table}

\subsection{ Experiments on Advanced Classifier architectures}
We empower UDG with wider (WRN28) and deeper (DenseNet) classifier.
Table~\ref{tab:arch} shows the comparison results with \cifarh as In-D samples using WRN28 and DenseNet architecture. 

As can be observed from the table, while UDG performs better on these architectures compared to ResNet18, it still lags far behind our results.

\begin{table}[t]
    \caption{
    Experiments on advanced model architectures.
    Performance comparison with UDG on \cifarh benchmarks. For our method, we use the results in the main paper with a ResNet18 classifier.
     We give advantage to UDG, which is reimplemented with deeper/wider WideResNet-28, DenseNet, while \sood's parameter number is equivalent to ResNet18.
    \textbf{Bold} are the best.
    }
    \centering
    \label{tab:arch}
    \scalebox{0.8}{
    \setlength{\tabcolsep}{3mm}{}
    \begin{tabular}{cccccc}
    \hline
        Architecture & OOD dataset & FPR@TPR95 ↓ & AUROC ↑ & AUPR In ↑ & AUPR Out ↑ \\ \hline
        \multirow{7}{*}{{\rotatebox{90}{\makecell[c]{WideResNet28\\UDG}}}}& 
        \svhn & 66.76 & 85.29 & 76.14 & 92.33 \\ 
        &  \cifar & 82.35 & 76.67 & 78.52 & 72.63 \\ 
        &  \scriptsize{\tinyimagenet} & 78.91 & 79.04 & 87.00 & 65.06 \\ 
        &  \texture & 73.62 & 79.01 & 85.53 & 67.08 \\ 
        &  \lsun & 77.04 & 79.79 & 87.49 & 66.93 \\ 
        &  \places & 72.25 & 81.49 & 66.72 & 90.65 \\ \cline{2-6}
        & \textbf{Mean}$\scriptstyle{\pm Std}$ & 75.16$\scriptstyle{\pm 5.49}$ & 80.22$\scriptstyle{\pm 2.93}$ & 80.23$\scriptstyle{\pm 8.11}$ & 75.78$\scriptstyle{\pm 12.44}$ \\ \hline
        
          \multirow{7}{*}{{\rotatebox{90}{\makecell[c]{DenseNet\\UDG}}}}
          & ~ \svhn & 80.67 & 75.54 & 75.65 & 70.99  \\ 
        ~ &   \cifar & 85.87 & 74.06 & 77.16 & 68.90  \\ 
        ~ &  \tinyimagenet & 82.36 & 76.81 & 85.76 & 61.56  \\ 
        ~ &  \texture & 76.32 & 78.93 & 63.79 & 89.02  \\ 
        ~ &  \lsun & 79.12 & 78.91 & 66.83 & 88.23  \\ 
        ~ &  \places & 73.59 & 76.27 & 82.76 & 65.20  \\ \cline{2-6}
        ~ &  \textbf{Mean}$\scriptstyle{\pm Std}$ & 79.66$\scriptstyle{\pm 4.36}$ & 76.75$\scriptstyle{\pm 1.92}$ & 75.33$\scriptstyle{\pm 8.64}$ & 73.98$\scriptstyle{\pm 11.79}$ \\ \hline
        \multirow{7}{*}{\rotatebox{90}{\textbf{\makecell[c]{\sood \\ (Ours, Res18)}}}} & \svhn  & \textbf{51.6} & \textbf{88.99} & \textbf{80.89} & \textbf{94.81} \\ 
        &  \cifar  & \textbf{50.17} & \textbf{87.76} & \textbf{88.18} & \textbf{87.79} \\ 
        &  \scriptsize{\tinyimagenet}  & \textbf{46.07} & \textbf{89.42} & \textbf{89.73} & \textbf{89.28 }\\ 
        &  \texture  & \textbf{42.22 }& \textbf{90.56} & \textbf{94.43} & \textbf{85.13} \\ 
        &  \lsun  & \textbf{47.85} & \textbf{89.96} & \textbf{90.33} & \textbf{89.23} \\ 
        &  \places  & \textbf{47.72} & \textbf{89.3} & \textbf{74.83} & \textbf{96.48} \\ \cline{2-6}
        &  \textbf{Mean}$\scriptstyle{\pm Std}$ & \textbf{47.61}$\scriptstyle{\pm 3.29}$ & \textbf{89.33}$\scriptstyle{\pm 9.95}$& \textbf{86.4}$\scriptstyle{\pm 7.19}$ & \textbf{90.45}$\scriptstyle{\pm 4.33}$\\
         \hline
    \end{tabular}
    }
\end{table}

\section{Qualitative Results}
In this section, we demonstrate several batches of visual examples of \sood including both In-D and OOD cases.

\subsubsection{\textit{In-D samples with their syntheses.}} 
\begin{figure}[tpb]
	\centering\includegraphics[width=0.9\linewidth]{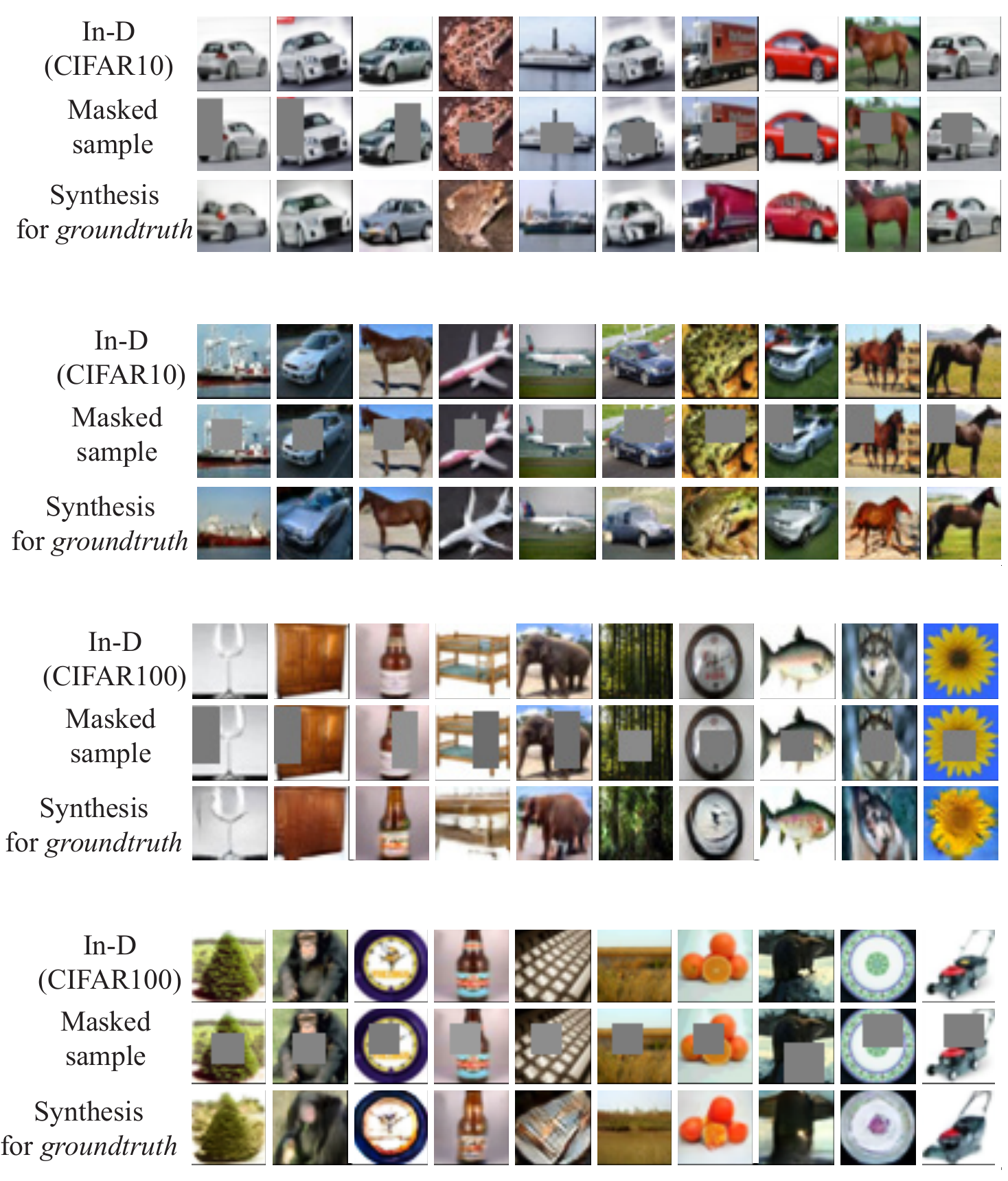}
	\caption{Visualization results of \sood with \cifar / \cifarh as In-D. We exemplify several In-D samples in each panel's first row, following the intermediate masked version, and the last row presents their corresponding synthetic version generated by \sood with the groundtruth labels.}
	\label{fig:Indcifar10}
\end{figure}

Fig.~\ref{fig:Indcifar10}
visualizes In-D samples and their corresponding
 syntheses from \cifar and \cifarh, respectively. Note that we expect the syntheses to resemble the input images for In-D samples with correct labels.


\subsubsection{\textit{OOD samples with their syntheses.}} 
\begin{figure}[tpb]
	\centering\includegraphics[width=1\linewidth]{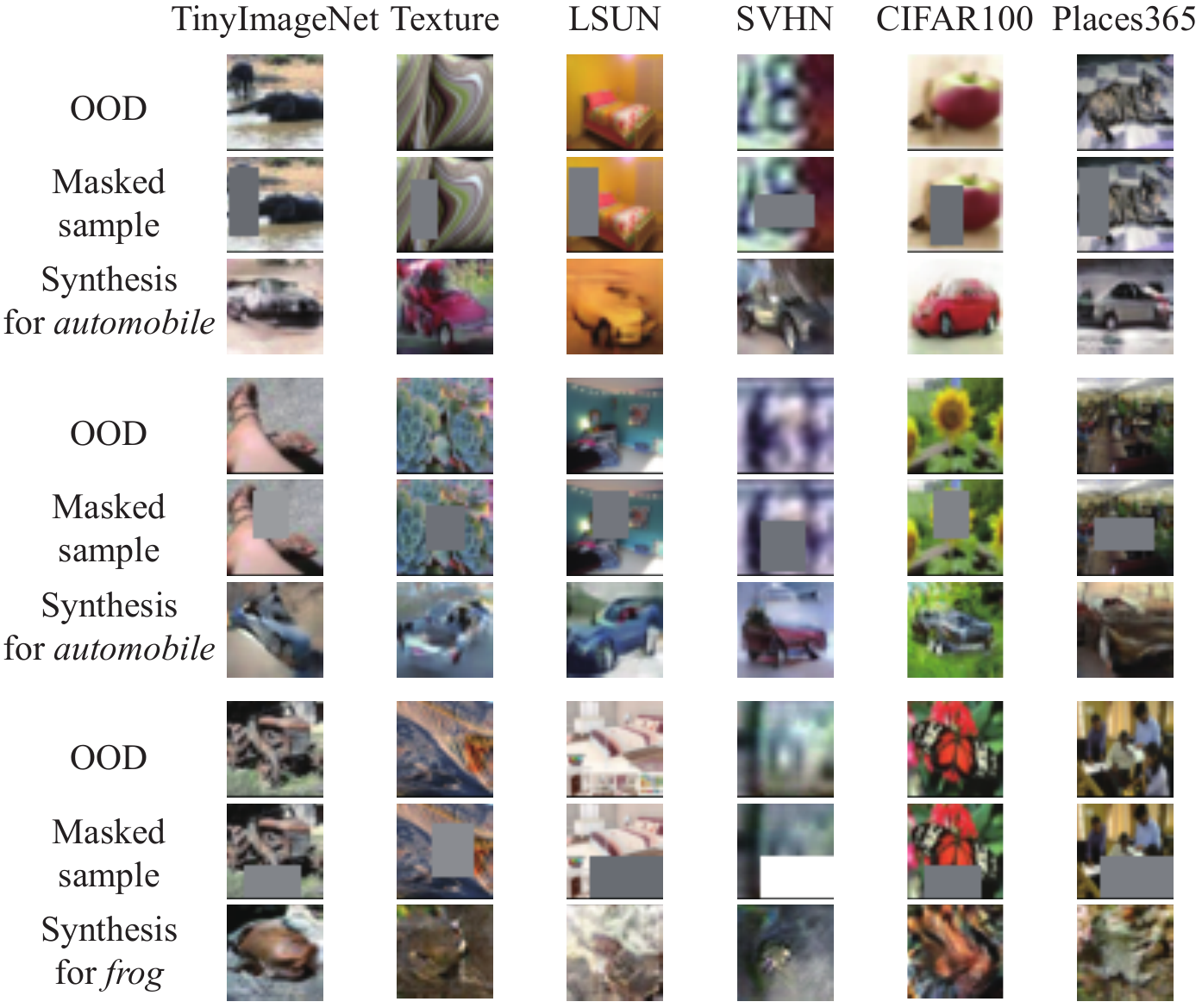}
	\caption{ OOD visualization results of \sood trained on \cifar. 
	In each panel, we exemplify OOD samples across six OOD datasets in the first row, following is the intermediate masked version, the last row presents their corresponding synthetic version generated by \sood with the given semantic label, the same below.}
	\label{fig:OODcifar10}
\end{figure}

\begin{figure}[tpb]
	\centering\includegraphics[width=1\linewidth]{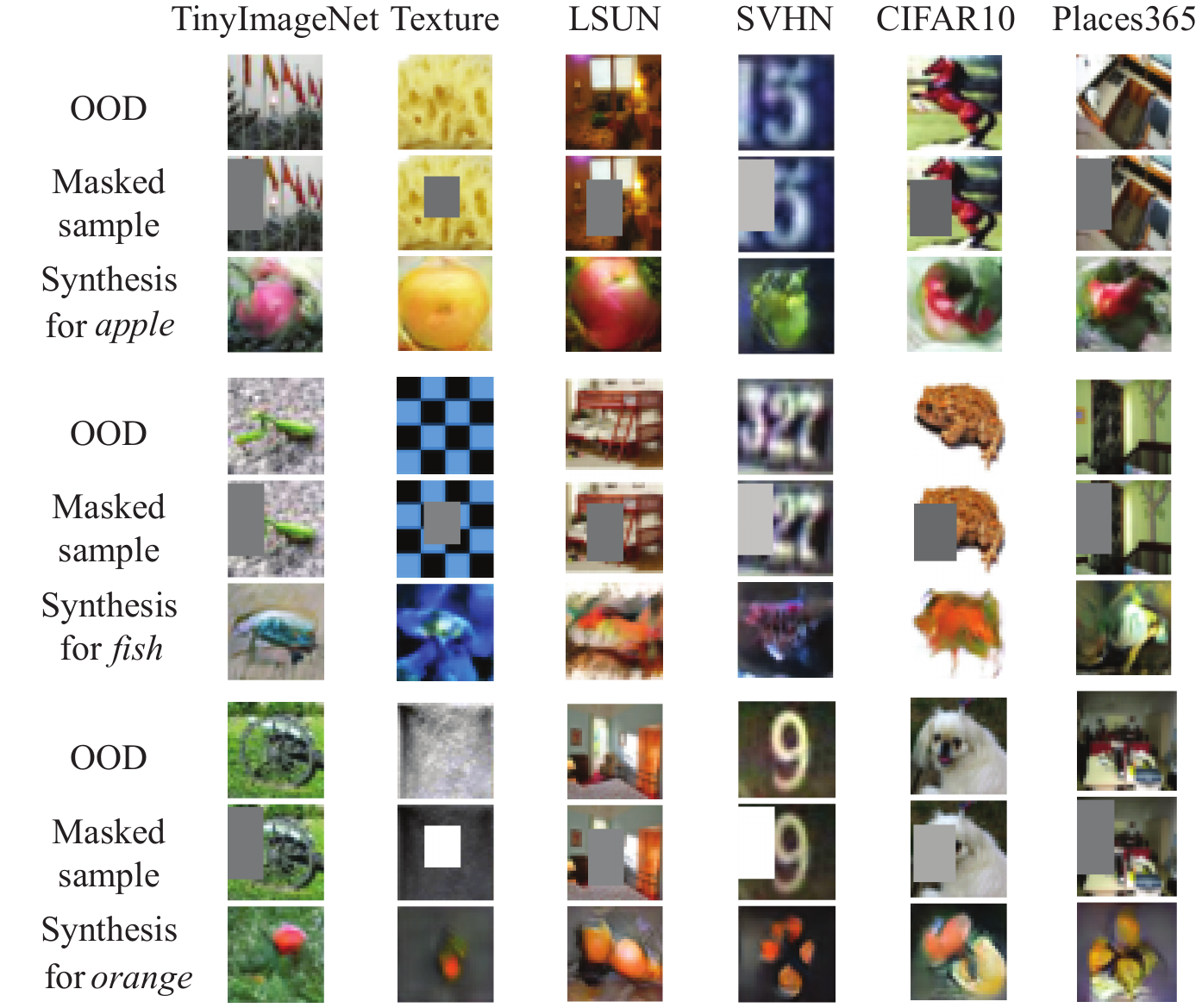}
	\caption{ OOD visualization results of \sood trained on \cifarh. }
	\label{fig:OODcifar100}
\end{figure}

Fig.~\ref{fig:OODcifar10} visualizes OOD samples from six datasets, which are employed in the \cifar benchmarks, along with their corresponding masked images and the syntheses generated by our \sood. 
In Fig.~\ref{fig:OODcifar100}, the In-D dataset changes to \cifarh. We employed OOD samples sourced from the same six OOD datasets as those of the \cifarh benchmarks in Fig.~\ref{fig:OODcifar100}.   Note that when OOD is fed to \sood, we prefer to have a clear distinction between the synthesis generated by \sood and the input image.

\section{Further Discussion}
\subsection{Computational Cost Analysis}
\sood is designed as an auxiliary model that works in parallel with the classifier. This auxiliary architecture ensures \sood  
 a plug-and-play model without compromising the classifier's accuracy. Meanwhile, \sood can satisfy high-performance requirements in the context of OOD detection. However, as an auxiliary model, \sood inevitably introduces extra computation and memory costs. 
 
Table~\ref{tab:cost} summarizes the computational cost of \sood, 
and that of ODIN, i.e., ResNet18, 
and that of widely adopted classifier architectures, ResNet18, WResNet28, WResNet101 in terms of number of multiply add operations (MAC), and number of model parameters (Params).
As can be observed, the cost of basic version of \sood, i.e. $\mathcal{C}_{b}$, \E, \D, is relatively small, Params 4.552M, MACs 0.408G, when compared to that of ResNet18 (Params 11.174M, MAC 0.556G) and other widely adopted classifier architectures, e.g., WResNet28 (Params 36.479M, MACs 5.248G). Note that the performance of basic \sood, whose anomalous scoring model only contains $\mathcal{C}_b$, is still acceptable as shown in Table~\ref{tab:ablation_cifar10}  and Table~\ref{tab:ablation_cifar100}. Thus, if computational cost is a real concern in practice, the operator can adopt \sood with $\mathcal{C}_b$ alone as an anomalous scorer.
For the \sood supported by IQA models, e.g., LPIPS, DISTS, the total computational cost is comparable to that of WResNet28 or WResNet101, yet slightly larger than ResNet18.
Thus, if the detection ability is put at the first place, one can explore to enhance the anomalous scoring model by employing extra IQA models.  Actually, there is a trade-off between the OOD detection performance and the computational cost of \sood, and our anomalous scoring model leaves the design space for the deployer to explore according to the real-world application.  
 
\begin{table}[H]
	\centering
	\caption{Computational and memory costs of \sood and its components. }
\label{tab:cost}
\fontsize{6}{9}\selectfont
\begin{tabularx}{\textwidth}{ P{1.8cm} P{0.5cm} s P{0.5cm} s s s s s s  }
\toprule 
\begin{tabular}{@{}c@{}}\bf{Model} \end{tabular}
& \begin{tabular}{@{}c@{}} \bf{E} \end{tabular} 
& \begin{tabular}{@{}c@{}} \bf{D} \end{tabular}
& \begin{tabular}{@{}c@{}} \bf{$\mathcal{C}_{b}$} \end{tabular}
&  \begin{tabular}{@{}c@{}} \bf{\textsc{MoodCat}} \\ \bf{basic} \end{tabular}
&  \begin{tabular}{@{}c@{}} \bf{LPIP} \\ \bf{/DISTS} \end{tabular}
&  \begin{tabular}{@{}c@{}} \bf{\textsc{MoodCat}} \end{tabular}
&  \begin{tabular}{@{}c@{}} \bf{ResNet}\\\bf{18} \end{tabular}
&  \begin{tabular}{@{}c@{}} \bf{WResNet}\\\bf{28} \end{tabular}
&  \begin{tabular}{@{}c@{}} \bf{WResNet} \\\bf{101}\end{tabular}\\
	\midrule
	\textbf{Params (M)} & 0.460   & 3.821   & 0.271 & \textbf{4.552} & 14.715   & 33.982 & 11.174 &36.479 & 126.89\\ 
	\textbf{MACs (G)}   & 0.0049  & 0.297   & 0.105 & \textbf{0.408}  & 0.630   & 1.718  & 0.556  &5.248 & 22.84 \\ \bottomrule
\end{tabularx}
\end{table}
\subsection{Failure Cases}
\begin{figure}[tpb]
	\centering\includegraphics[width=\linewidth]{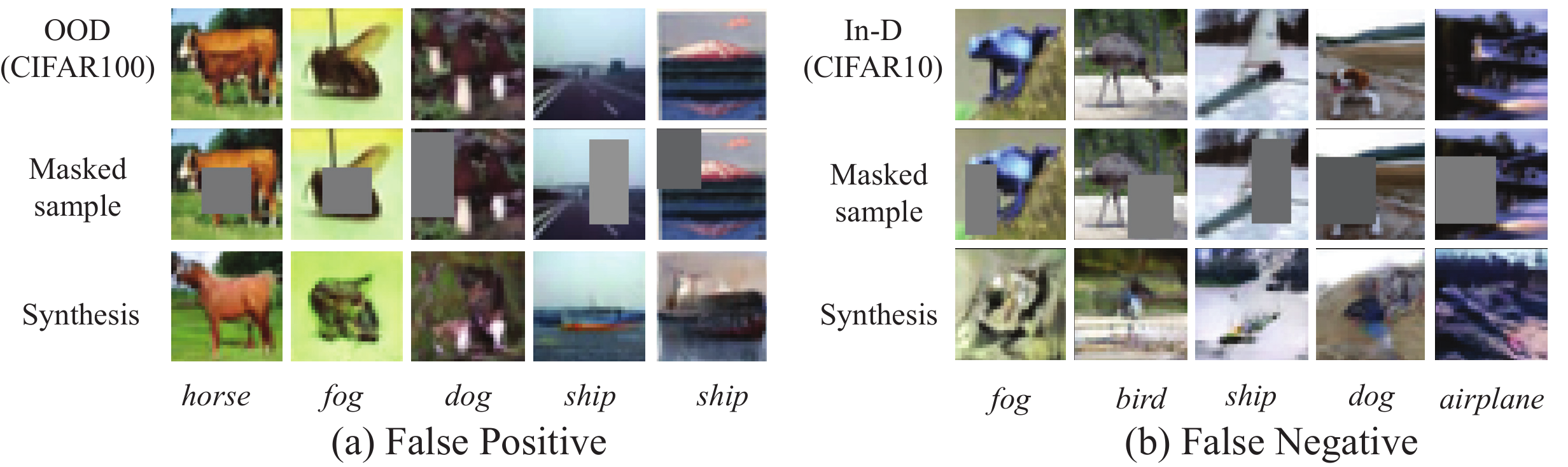}
	\caption{Failure cases of \sood. We exemplify both False Positive   and False Negative failure cases in (a) and (b), respectively. (a) False Positive failure cases, where samples come from OOD dataset (\cifarh) are falsely identified as In-D samples (\cifar). (b) False Negative failure cases, where samples belong to In-D are wrongly flagged as OOD samples. The predicted label for each input sample are provided under the corresponding synthetic image. }
	\label{fig:failure case}
\end{figure}

Fig.~\ref{fig:failure case} demonstrates some of \sood's failure cases.
In Fig.~\ref{fig:failure case} (a), the OOD samples sourcing from \cifarh, are falsely distinguished as In-D samples (\cifar). As can be seen, OODs and their synthetic images resemble to each other for same degree. For example, the first column's ``cattle'' partly contains some features such as legs and the tail, which match the given semantic label ``horse'' well, resulting in the synthesis having high image quality while resembling to the input image, therefore leading to the final misjudgement.

Fig.~\ref{fig:failure case} (b) presents several False Negative samples, i.e., samples sourcing from In-D are wrongly predicted as OOD samples. As can be observed, the In-D sample with rare characteristics, e.g. a blue fog, an ostrich with its head down, are more likely to be misclassified as OOD. In addition, if the mask happens to cover the object completely, \sood can hardly recover the input image without necessary features, as the cases shown in the third and fourth columns of Fig.~\ref{fig:failure case} (b). Moreover, an poor semantic meaning in the In-D sample itself can lead to the final misclassification. For example, in the last column of Fig.~\ref{fig:failure case} (b), even humans can hardly tell what is depicted in the input image, let alone \sood. 

\end{document}